\documentclass[10pt, a4paper, twocolumn, teaser, showabstract]{naverlabseurope}

\usepackage{lipsum}
\usepackage{multicol}
\usepackage{tikz,tkz-kiviat,pgfplots}

\usepackage[dvipsnames]{xcolor}
\usepackage{booktabs} %
\usepackage{tabularx}
\usepackage{graphicx}%
\usepackage{stmaryrd} %
\usepackage{bm}
\usepackage{subcaption}
\usepackage{multirow}
\usepackage{stfloats}  %

\DeclareMathOperator{\trace}{Tr}
\DeclareMathOperator{\diag}{diag}
\DeclareMathOperator{\sigmoid}{sigmoid}
\DeclareMathOperator*{\argmax}{argmax}

\definecolor{cvprblue}{rgb}{0.21,0.49,0.74}

\makeatletter
\renewcommand{\paragraph}{%
  \@startsection{paragraph}{4}%
  {\z@}{0.25ex \@plus 1ex \@minus .2ex}{-1em}%
  {\normalfont\normalsize\bfseries}%
}
\makeatother

\expandafter\def\expandafter\normalsize\expandafter{%
    \normalsize%
    \setlength\abovedisplayskip{3pt}%
    \setlength\belowdisplayskip{3pt}%
    \setlength\abovedisplayshortskip{3pt}%
    \setlength\belowdisplayshortskip{3pt}%
}

\usepackage[hang,flushmargin]{footmisc}

\newcommand{\ours}{CondiMen\xspace}
\newcommand{\Ours}{\ours}

\graphicspath{{figures/}}

\title{\Ours: Conditional Multi-Person Mesh Recovery}
\titlerunning{\Ours: Conditional Multi-Person Mesh Recovery}

\correspondingauthor{romain.bregier@naverlabs.com\\
Salma Galaaoui is now at Valeo AI.}

\authors{Romain Brégier$^{\star}$ \authsep Fabien Baradel$^{\star}$ \authsep Thomas Lucas$^{\star}$ \authsep Salma Galaaoui$^{\star\dagger}$ \\ \authsep Matthieu Armando$^{\star}$ \authsep Philippe Weinzaepfel$^{\star}$ \authsep Grégory Rogez$^{\star}$}
\affiliations{$^{\star}$NAVER LABS Europe \\$^{\dagger}$LIGM, Ecole des Ponts, Univ Gustave Eiffel, CNRS, Marne-la-Vallee, France}
\website{}
\websiteref{}
\contributions{}

\teaserfig{
  \centering
  \vspace{4pt}
  \includegraphics[width=\textwidth]{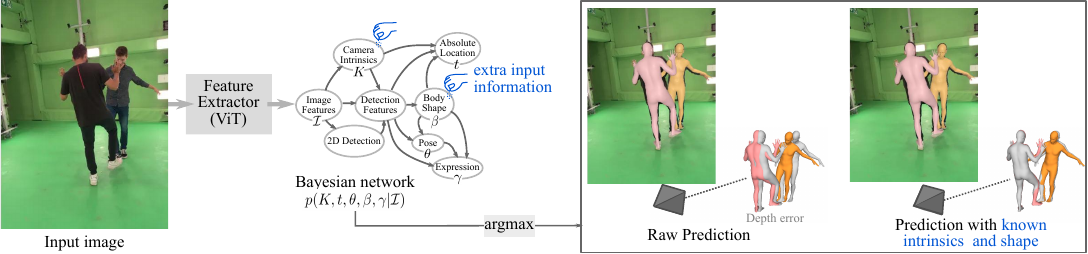}\\
  \vspace{-4pt}
}
  
\teasercaption{\label{fig:overview}
\textbf{
        CondiMen: A Recipe for HMR.
        }
        Recovering 3D human meshes from an image is challenging as predictions that look plausible in 2D can be inaccurate in 3D (ground-truth meshes in gray).
        To improve predictions, we propose a method leveraging additional information that may be available -- such as camera calibration, body shape, distance to the camera, or multi-view observations.
       We decompose mesh recovery into a human detection and attribute estimation problem, modeling the joint probability of these attributes (pose, body shape, \etc{}) using a Bayesian network.
        It enables efficient inference, compatible with 20FPS real-time applications.
}

\begin{document}

\begin{abstract}
 Multi-person human mesh recovery (HMR) consists in detecting all individuals in a given input image, and  predicting the body shape, pose, and 3D location for each detected person. 
The dominant approaches to this task rely on neural networks trained to output a single prediction for each detected individual.
In contrast, we propose \Ours, a method that outputs a joint parametric distribution over likely poses, body shapes, intrinsics and distances to the camera, using a Bayesian network.
This approach offers several advantages.
First, a probability distribution can handle %
some inherent ambiguities of this task -- such as the uncertainty between a person's size and their distance to the camera, or more generally the loss of information that occurs when projecting 3D data onto a 2D image.
Second, the output distribution can be combined with additional information to produce better predictions, by using \eg{} known camera or body shape parameters, or by exploiting multi-view observations.
Third, one can efficiently extract the most likely predictions from this output distribution, making the proposed approach suitable for real-time applications.
Empirically we find that our model i) achieves performance on par with or better than the state-of-the-art, ii) captures uncertainties and correlations inherent in pose estimation and iii) can exploit %
additional information at test time, such as multi-view consistency or body shape priors.
\Ours spices up the modeling of ambiguity, using just the right ingredients on hand.
\end{abstract}

\maketitle

\section{Introduction}

Recovering people characteristics in 3D from images is essential for a variety of applications, ranging from human behavior analysis to robotic systems in crowded environments. In this work, we present a method that detects individuals in images and then predicts 3D meshes, encoding their pose, body shape, and location.

Human mesh recovery is an ill-posed problem, as %
different meshes could be plausible for a single input image due to factors like clothing, occlusions, and the projective nature of 2D imaging.
In particular, the apparent 2D size of a person in an image depends on their actual size in 3D, the distance to the camera, and the camera's focal length (see supp.\ mat.\ for examples). 
Despite this uncertainty, state-of-the-art methods~\cite{multi-hmr2024, romp, bev, kanazawa2018hmr} typically predict deterministic outputs. These methods are optimized to minimize an empirical loss over the training data, and in the presence of ambiguity, they tend to predict average attributes -- those that occur most frequently in the training data -- resulting in a loss of accuracy when faced with aleatoric uncertainty.

A probabilistic framework offers an elegant way to handle such uncertainty.
Methods that predict probability distributions -- allowing for the sampling of hypotheses~\cite{li2019generating, kolotouros2021probabilistic, muller2024buddi}  -- 
or that provide confidence estimates~\cite{sidenbladh2000stochastic, kolotouros2021probabilistic, dwivedi2023poco} have been proposed in previous work. %
However, most of them focus solely on relative pose estimation and fail to consider the body shape or 3D location, which are critical in many applications.
Moreover, there are scenarios where additional information such as camera calibration, body shapes previously scanned, or multi-view image captures are available and could provide useful cues to improve predictions.
Incorporating such external inputs into existing methods is unfortunately challenging and often requires iterative optimization techniques~\cite{kolotouros2021probabilistic}, limiting their practical application.

In this paper, we propose a solution to this issue by treating mesh recovery as a task of jointly predicting various attributes (pose, body shape, location, \etc), and by regressing a joint probability distribution over these attributes.
Specifically, we introduce a parametric Bayesian model~\cite{bishop2006pattern} illustrated in~Fig.~\ref{fig:overview} that, given an input image, outputs a joint probability distribution over camera intrinsics, human detections, poses, body shapes and 3D locations.
This probabilistic formulation accounts for ambiguities of the multi-person mesh recovery tasks and allows for the efficient use of external information during inference.
Additionally, multi-view images can seamlessly be used within our framework to improve 3D mesh recovery -- something which is not straightforward with monocular deterministic methods.
This ability to make multi-view predictions at test time, despite being trained on monocular data only is advantageous because high-quality, diverse monocular data is far more abundant and easier to collect than multi-view data, which often requires complex setups.

For our experiments, we train a model from synthetic-only data, using BEDLAM~\cite{black2023bedlam} as well as images we generated to further increase data variability.
We demonstrate that the Bayesian nature of our model offers flexibility and the ability to exploit available external knowledge, at test time.
For instance, we show that leveraging known ground-truth quantities (\eg camera intrinsics or body shape) within the model can improve predictions in a zero-shot manner.
It can also be used to exploit consistency across multiple input images, for instance by enforcing a constant body shape for a given person.
We validate the performance of our approach against existing state-of-the-art methods on monocular and multi-view datasets~\cite{3dpw, mupots, hi4d, h36m, rich}, showing competitive results and the ability to handle uncertainty effectively.

\section{Related work}

\paragraph{Human Mesh Recovery from a Single Image.}
The development of parametric 3D models~\cite{SMPL:2015,xu2020ghum,pavlakos2019expressive} has advanced the field of human pose estimation from 3D skeleton regression ~\cite{rogez2017lcr,dope,chen2016synthesizing} to the prediction of full human body meshes~\cite{kanazawa2018hmr,goel2023humans4d,kolotouros2019spin,zhang2021pymaf,li2022cliff,joo2021exemplar,dwivedi2024tokenhmr,sarandi2024neural}.
More recently, research has focused on estimating expressive human meshes~\cite{black2023bedlam,sun2024aios,multi-hmr2024,patel2021agora,hewitt2024look} and placing these meshes within real-world coordinate systems~\cite{wang2025tram,shin2024wham}.
The pioneering work by \citet{kanazawa2018hmr} introduces a method to predict SMPL~\cite{SMPL:2015} parameters, along with weak-perspective re-projection parameters, from a single cropped image of a person.
Subsequent methods have improved on this approach, either by refining the network architecture~\cite{goel2023humans4d,zhang2021pymaf,li2022cliff} or by leveraging new training data and protocols~\cite{patel2021agora,black2023bedlam}.
Recent advances have also enabled whole-body pose estimation, \ie, including facial expression and hand poses~\cite{ehf,osx,pixie,hand4whole,choutas2020expose,rong2021frankmocap,zhou2021monocular,pymafx2023,cai2023smplerx,hewitt2024look}.
A few methods now tackle the detection and regression of multiple human meshes within a single network~\cite{romp,bev,psvt,multi-hmr2024,sun2024aios}, though these approaches typically produce deterministic outputs.
In contrast, our proposed Bayesian network for multi-person whole-body mesh recovery outputs a probability distribution, allowing it to handle ambiguities, integrate external information, and %
fuse multi-view predictions
for more robust performance in complex scenarios.

\paragraph{Multi-view Human Mesh Recovery}
A first category of methods to tackle multi-view mesh recovery assumes known camera calibration to extend single view reconstruction to multi-view settings~\cite{bogo2016keep,zhu2024muc,zhang2023probabilistic,davoodnia2025upose3d}.
In particular, SMPLify-X~\cite{bogo2016keep} performs 3D reconstruction in a unified coordinate system by iteratively minimizing 2D keypoint reprojection errors.
Some learning-based methods also follow this  %
calibrated approach~\cite{tu2020voxelpose,ye2022faster,iskakov2019learnable,choudhury2023tempo}.
For example, \citet{iskakov2019learnable} propose a learnable triangulation solution.
A second category of methods addresses the uncalibrated setup, where predictions from multiple views are used to estimate camera parameters and merged either through handcrafted averaging~\cite{pavlakos2022human,yu2022multiview,li20213d} or learning-based techniques~\cite{qiu2019cross,zhu2024muc}. 
Recent adaptive frameworks can handle both calibrated and uncalibrated settings but are still limited to single-person~\cite{jia2023delving,xie2024visibility}.
In contrast, our approach supports multi-person mesh recovery across multiple views.

\paragraph{Probabilistic Human 3D Pose.}
Early methods addressed pose uncertainty %
through optimization-based formulations~\cite{sidenbladh2000stochastic,sminchisescu2001covariance} or by predicting multiple 3D poses from 2D cues~\cite{li2019generating,zhang2022uncertainty,jahangiri2017generating}.
\citet{li2019generating} use a Mixture Density Network to infer a distribution of 3D joints from 2D joints, while \citet{zhang2022uncertainty} model separate Gaussian distributions for 2D keypoints and depth.
Recent works leverage Normalizing Flow (NF) to %
model pose or shape probability distributions~\cite{li2021human,kolotouros2021probabilistic,sengupta2023humaniflow, wehrbein2021probabilistic}, with \citet{kolotouros2021probabilistic} conditioning NF on image features to predict SMPL parameters.
Another line of research leverages denoising diffusion models~\cite{yuan2023physdiff,huang2024closely,muller2024buddi} that can account for uncertainties in depth, body shape, and camera intrinsics. %
Extracting the most likely prediction with such approach is often computationally intensive however.
Other strategies include using multiple prediction heads with a \emph{best-of-N} loss~\cite{biggs20203d}, estimating Gaussian distributions over SMPL parameters~\cite{sengupta2021probabilistic}, or quantizing human mesh representations~\cite{fiche2023vq}.
\citet{sengupta2021hierarchical} use a hierarchical matrix-Fisher distribution for each SMPL rotation parameters following the kinematic tree.
In contrast, we introduce a Bayesian network head that can handle uncertainty while allowing for efficient inference, use of external information, and multi-view predictions in a simple and elegant way.

\section{Method}

We present our method for detecting people and estimating their 3D whole-body attributes from a single input image.
This method called \emph{\ours} (short for \emph{conditional multi-person mesh recovery}) is illustrated in Fig.~\ref{fig:overview}.
We extract image features %
using a Vision Transformer (ViT) backbone~\cite{dosovitskiy2020image}, 
which we use as conditioning variables in a joint probability distribution that models people appearing in the image with their different attributes (3D location, body shape, \etc{}).
We model this joint distribution as a trained Bayesian network.
At inference, we efficiently detect humans and predict their attributes, by sequentially extracting modes of the conditional distributions. %
This probabilistic framework allows us to leverage %
different information available at inference, such as known %
camera intrinsics or specific body shapes %
to enhance prediction accuracy.

\subsection{Problem formulation}

\paragraph{Human parametrization.}
To encode human meshes, we rely on a parametric body model, namely SMPL-X~\citep{choutas2020expose}. SMPL-X provides a whole-body parametrization decoupled into an absolute 3D location $t$, a list of bone orientations $\theta$ modeling the pose, a vector $\beta$ modeling the body shape and a vector $\gamma$ modeling the facial expression.

\paragraph{Bayesian network.}
We model the multi-person mesh recovery problem as a probabilistic optimization problem. 
Given some input image features $\mathcal{I}$, we aim at predicting the value of different random variables:
the intrinsic parameters $K$ of the camera, as well as attributes of people visible in the image $(t, \theta, \beta, \gamma)$\footnote{We describe the single-person case here to simplify notations.}.
We consider the joint probability distribution of these variables conditioned on image features, and aim to extract the most likely prediction:
\begin{equation}
\hat{K}, \hat{t}, \hat{\theta}, \hat{\beta}, \hat{\gamma} = \argmax ~ p(K, t, \theta, \beta, \gamma \vert \mathcal{I}),
\label{eq:principal_mode}
\end{equation}
where $p(K, t, \theta, \beta, \gamma \vert \mathcal{I})$ denotes the associated probability density.
While conceptually appealing, modeling such joint distribution from limited data is challenging due to the high dimension of the representation space ($\dim(K){=}3$, $\dim(t){=}3$, $\dim(\beta){=}11$, $\dim(d){=}1$, $\dim(\theta){=}53{\times}3$, $\dim(\gamma){=}10$ in our setting).

A classical solution to this \emph{curse of dimensionality} is to adopt the naive Bayes assumption, which posits that different variables are conditionally independent given the image features. This results in a factorized form: %
$p(K, t, \theta, \beta, \gamma \vert \mathcal{I}) \approx p(K \vert \mathcal{I}) \cdot p(t \vert \mathcal{I}) \cdot p(\theta \vert \mathcal{I}) \cdot p(\beta \vert \mathcal{I}) \cdot p(\gamma \vert \mathcal{I})
$,
where $p(x \vert y)$ denotes probability density at $x$ conditioned on the value $y$.
In practice, each conditional density $p( \cdot \vert \mathcal{I})$ is generally assumed to belong to a parametric family, with parameters that are functions of the input $\mathcal{I}$, \eg regressed using a neural network.
Multi-HMR~\cite{multi-hmr2024} and most other deterministic methods~\cite{bev,romp,psvt} can be thought of as special cases within this framework. They use regression objectives equivalent to a naive Bayes formulation, assuming probability distributions with constant dispersion terms.
However, a key limitation of the naive Bayes assumption is that it ignores inter-variable dependencies. These dependencies are often crucial for scene understanding:
for instance a small person A appearing the same size in 2D as a taller person B is likely to be closer to the camera than B, other things being equal.

To address these challenges, we adopt a relaxed hypothesis by modeling the joint distribution as a Bayesian network, decoupling variables into a directed acyclic graph of conditional distributions, illustrated in Fig.~\ref{fig:overview}.
The ability of deep neural networks to model complex, high-dimensional data in an auto-regressive manner has been demonstrated %
across various domains including text~\cite{van2016conditional,brown2020language} and images~\cite{esser2021taming}. 
Our Bayesian model is inspired by auto-regressive approaches, encoding relationships between variables in a cascaded fashion.

\subsection{Conditional distributions}

\begin{figure}
    \small
    \centering
    \includegraphics[width=\linewidth]{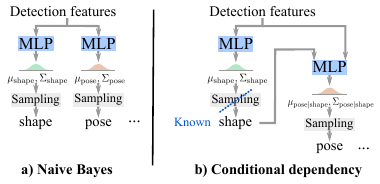}
    \caption{\label{fig:conditional_dependency_modeling}
    \textbf{Modeling conditional dependency}.
    We predict probability distributions for different human attributes (e.g., pose, shape, distance) and efficiently sample the most likely predictions.
    Rather than treating each attribute independently (a), we capture their interdependency by modeling conditional distributions (b), leading to more coherent results.
    Our framework can also incorporate additional input information when available (such as a known shape attribute, shown in blue), to further improve prediction accuracy.
    }
    \vspace{-0.1cm}
\end{figure}

We implement our Bayesian network using a cascade of Multi-Layer Perceptrons (MLP). %
Each MLP outputs parameters of a probability distribution associated with a random variable of our mesh recovery problem, such as the pose of a detected person. This distribution is conditioned on the value of its parent variables in the Bayesian network's graphical model%
, which allows us to, for example, encode the likelihood of a body pose given a specific body shape and image detection features
(see Fig.~\ref{fig:overview} and Fig.~\ref{fig:conditional_dependency_modeling}).
In this section, we present the variables and distributions considered in our experiments.

\paragraph{Camera intrinsics.}
We assume a pinhole camera model with focal length $f{>}0$ and principal point $p \in \mathbb{R}^2$. We predict the parameters of Gaussian distributions for $\ln(f)$ and $p$ %
, conditioned on image features, specifically on the [CLS] token output by the ViT image encoder. Considering $\ln(f)$ instead of $f$ %
ensures that $f>0$, and is mathematically equivalent 
to modeling $f$ with a log-normal distribution.

\paragraph{2D detection.}
Image features produced by the ViT encoder consist of patch tokens $P_{u,v}$ defined along a 2D regular grid $G= \left\lbrace (u,v) \right\rbrace_{u=1\dots w, v=1 \dots h}$.
We encode people detections as binary variables $s_{u,v}$ along this grid, which indicate whether a reference keypoint of a person projects into a grid cell $(u,v) \in G$, following the CenterNet object detection framework~\cite{centernet}.
For each variable, we predict a score $p(s_{u,v} \vert \mathcal{I})$ encoding the detection likelihood, which is regressed from the corresponding patch features $P_{u,v}$. In practice, we use the person's head as reference keypoint (see Fig.~\ref{fig:results_examples}a), assuming that at most one person is detected in each grid cell.
At inference, detection is performed via score thresholding and local non-maxima suppression.

\paragraph{Detection features.}
For each detected person, we consider a latent variable consisting of image patch features $P_{u,v}$ of the detected grid cell $(u,v)$ augmented with camera ray embeddings, following an approach similar to Multi-HMR~\cite{multi-hmr2024}. These \emph{detection features} thus depend on camera intrinsics, and will serve as main conditioning variable for predicting the different human attributes.

\paragraph{Human attributes.}
SMPL-X parameterizes body shape and expressions as latent vectors of a PCA space of dimension $D$ ($D{=}11$ for shape, $D{=}10$ for expression, in our setup). We therefore model the conditional distributions for shape and expression as multivariate diagonal Gaussians.
The absolute 3D location of the person is decomposed into 2D coordinates $c$ of the reference keypoint in the image, and distance $d$ to the image plane.
To encode this distance, we use the  variable $\ln(d/f)$ referred to as \emph{encoded depth}, and model the conditional distributions for $c$ and $\ln(d/f)$ as normal distributions. The $\ln(d/f)$ encoding %
ensures that $d$ remains positive and allows for stronger conditioning on camera intrinsics. 
Lastly, pose is parameterized as a tuple $\theta \in SO(3)^{J}$ of $J=53$ bone orientations.
Since $SO(3)$ has a more complex topology than the PCA space of shapes and expressions, we model the conditional pose distribution as a product of independent matrix Fisher distributions $\Pi_{j=1}^{J} \mathcal{F}(\bm{F}_j)$
of density defined for a rotation matrix $\bm{R} \in SO(3)$ and some distribution parameters $\bm{F} \in \mathcal{M}_{3 \times 3}(\mathbb{R})$ by:
\begin{equation}
p_{\mathcal{F}(\bm{F})}(\bm{R}) = c(\bm{F}) \exp(\trace(\bm{F}^\top \bm{R})),
\label{eq:matrix_fisher}
\end{equation}
with $c(\bm{F})$ a normalization constant.

\paragraph{Distribution parametrization.}
\label{sec:distribution_parametrization}
To model a $D$-dimensional Gaussian distribution $\mathcal{N}(\bm{\mu}, \bm{\Sigma})$ while avoiding degenerate cases, we regress the mode $\bm{\mu} \in \mathbb{R}^{D}$ of the distribution along with dispersion parameters $\bm{\sigma} \in \mathbb{R}^{D}$, from which we construct the diagonal covariance matrix $\bm{\Sigma} = \diag(1 + \exp(\bm{\sigma}))^2$.
Similarly, we decompose the parameter $\bm{F} \in \mathcal{M}_{3 \times 3}(\mathbb{R})$ of a matrix Fisher distribution $\mathcal{F}(\bm{F})$ into a mode $\bm{R} \in SO(3)$ and dispersion parameters ($\bm{O} \in SO(3)$, $\bm{\Lambda} \in \mathbb{R}^{3}$), more suitable to be regressed by a MLP. These components are combined as follows:
$
    \bm{F} = \bm{R} \bm{O} \diag(\lambda \sigmoid(\bm{\Lambda}
    )) \bm{O}^\top,
$    
where $\sigmoid$ denotes the element-wise sigmoid function and $\lambda$ is a strictly positive scaling constant (we use $\lambda=2$ in our experiments).
Rotations are regressed as $3{\times}3$ matrices, which are orthonormalized using a differentiable special Procrustes operator implemented in RoMa~\citep{bregier2021deepregression}.

\paragraph{Normalization constant.}
The matrix Fisher probability density function of Eq.~\eqref{eq:matrix_fisher} is defined up to a constant $c(\bm{F})$. 
To evaluate this constant during training, we use numerical integration by sampling
36,864 rotations on a uniform SO(3) grid proposed by~\citet{yershova2010so3grid}.

\subsection{Inference}
\label{sec:inference}

At inference, we extract predictions from our Bayesian network in an efficient feed-forward manner.
Given a predicted distribution for a random variable (\eg{} body shape), we can sample a value for this variable and use it to predict conditioned distributions (\eg{} pose distribution, illustrated Fig.~\ref{fig:conditional_dependency_modeling}b). We iterate this process until all variables are sampled to generate hypotheses.

\paragraph{Mode extraction}
In practical applications, one is typically interested in extracting the most likely predictions given observations, \ie{} solutions of Eq.~\eqref{eq:principal_mode}.
Finding such solution is not straightforward however, notably due to the non-linearities introduced by the MLPs regressing conditional distributions parameters.
We therefore use a greedy but much more efficient alternative.
Similar to the sampling procedure described above, we proceed in a feed-forward iterative manner through the Bayesian graph. We assign to each variable a value corresponding to the mode of the associated conditional distribution, and we iterate until all variables are evaluated. 
This algorithm can be implemented very efficiently with the normal and Fisher distributions considered in our experiments, as mode values are readily available in the regressed distribution parameters.
Note that more advanced distributions such as normalizing flows~\cite{kolotouros2021probabilistic,sengupta2023humaniflow} could be used for greater expressivity, but these do not enable as efficient mode extraction as the simpler alternatives considered in this work.

\paragraph{Using known variables.}
A major benefit of modeling conditional distributions over deterministic regression is the ability to exploit additional information available. 
In many applications, prior knowledge such as camera intrinsic parameters (from calibration or image metadata), a person's body shape (when imaging a known individual), or their distance from the camera (using depth sensors, for instance) can be leveraged.
During inference, we simply inject the known values of corresponding variables into our Bayesian network instead of performing mode extraction, as shown Fig.~\ref{fig:conditional_dependency_modeling}b, to improve prediction consistency.

\begin{figure*}
\centering
\includegraphics[width=1.0\textwidth]{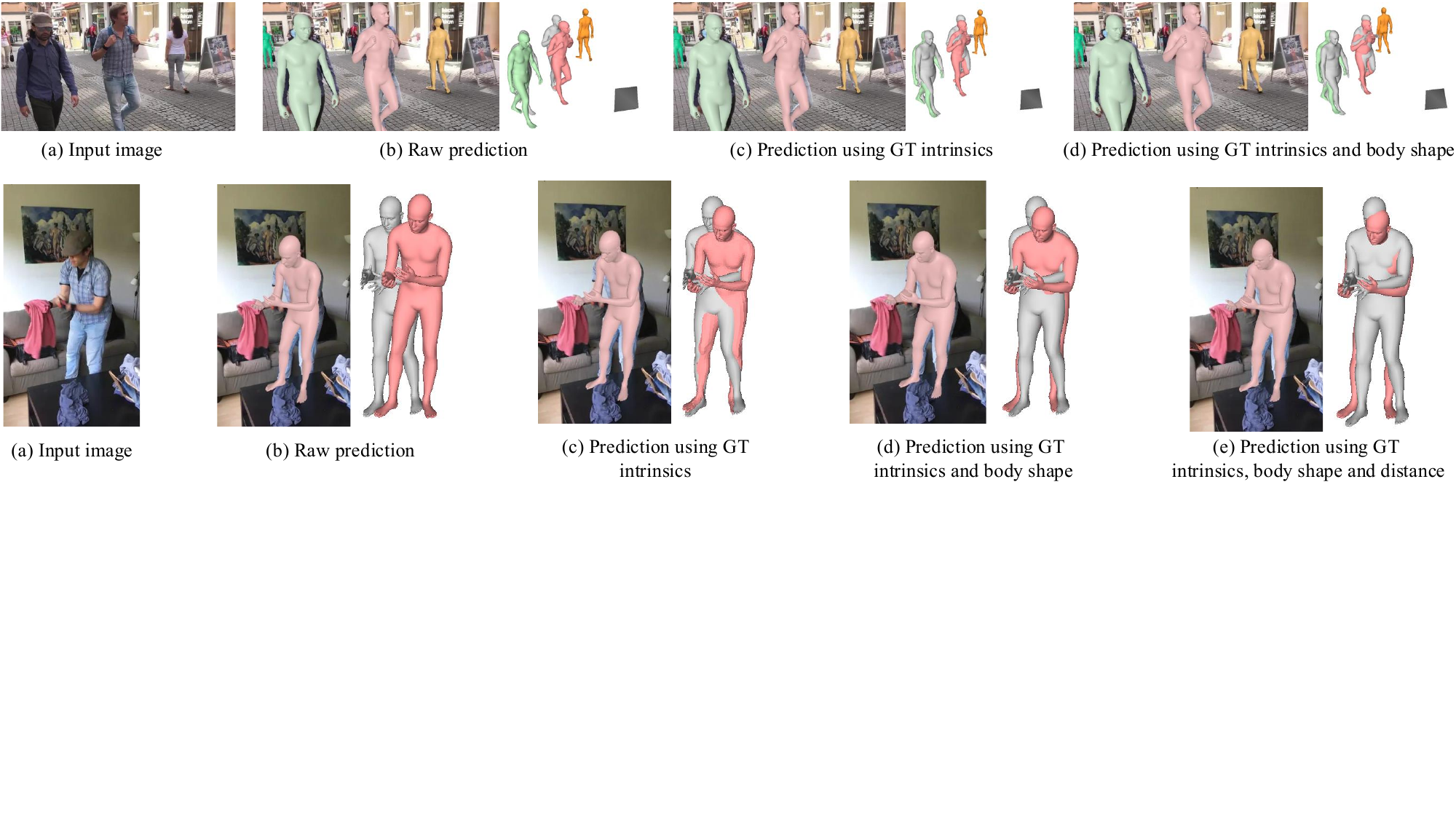}
\vspace{-4.6cm} %
\caption{\textbf{Qualitative results}.
Leveraging additional inputs, such as camera intrinsics and body shape, reduces errors and improves mesh accuracy. 
Ground-truth meshes are shown in grey for comparison.
\label{fig:3dpw_example1}
}
\end{figure*}

\paragraph{Multi-view prior.}
Better results can be obtained by exploiting $k$ simultaneous observations of the same person from different viewpoints, when available.
In our experiments, we assume that the camera poses are unknown.
We decompose the pose parameters into a global rigid orientation parameter $\theta_{0}$ and intrinsic, viewpoint-independent, pose parameters $\tilde{\theta}_j = \theta_{0}^{-1} \theta_j$, with $j=1 \dots J-1$. We then look for the multi-view prediction maximizing the product of posterior probabilities conditioned by image features:
\begin{equation}
 \prod_{i=1}^{k} p(K^{i}, t^i, (\theta_{0}^i, \theta_1 \dots \theta_{J-1}), \beta, \gamma | \mathcal{I}^i),
    \label{eq:likelihood_product}
\end{equation}
where variables specific to a view $i=1 \dots k$ are denoted with superscript $i$.
For greater efficiency, we also solve this problem greedily. %
We start by an initial rigid alignment of predictions (obtained separately for each view) to estimate global orientations $(\theta_{0}^i)_{i=1 \dots k}$, and then proceed
to find the optimal intrinsic orientations $\tilde{\theta}_j$. These are determined by minimizing
the product of Fisher probability densities $\prod_{i=1}^k p_{\mathcal{F}(\theta_{0}^i \bm{F}_{j}^i)}(\tilde{\theta_j})$
for each bone orientation  $j=1 \dots J-1$ and  view $i$
, see Eq.~\eqref{eq:matrix_fisher}.
This optimization admits a closed-form solution, which
consists for each bone orientation $\tilde{\theta}_j$ in the special Procrustes orthonormalization of $\sum_{i=1}^{k} \theta_{0}^i \bm{\bm{F}_{j}^i}$.

\paragraph{Matching.}
With multiple input views and multiple people per view, %
predictions from each view need to be matched to ensure  they correspond to the same person, before they can be combined.
For simplicity, we rely on Hungarian matching to obtain these matches, with cost matrices computed from pairs of %
single-view predictions after rigid alignment.

\subsection{Training}
\label{sec:training}
We train \ours to regress conditional distributions using an empirical  cross-entropy objective function denoted $\mathcal{L}_{\texttt{prob}}$. %
This objective function aims to
maximize the predicted log-probability density of the ground truth variables $(K^{*i}, s^{*i}, (t_j^{*i}, \theta_j^{*i}, \beta_j^{*i}, \gamma_j^{*i})_j)$ given images $i=1 \dots n$:
\begin{equation}
	\mathcal{L}_{\texttt{prob}} = - \cfrac{1}{n} \sum_{i=1}^n \log p(K^{*i}, s^{*i}, (t_j^{*i}, \theta_j^{*i}, \beta_j^{*i}, \gamma_j^{*i})_j \vert \mathcal{I}^i),
\end{equation}
with visible people indexed by $j$. This joint probability density  corresponds to the product of conditional probability densities predicted in our Bayesian network. We minimize the objective function by mini-batch gradient descent.

\paragraph{Mode guiding.}
To achieve better predictions, we found it beneficial to additionally guide the mode extraction procedure of our method.
At training time, we consider some input camera intrinsics $K$ and generate human attribute hypotheses using the procedure described in Sec.~\ref{sec:inference}.
This results in human mesh predictions $\hat{\mathrm{V}}$, composed of $|V|$ vertices centered at 3D location $\hat{t}$.
Denoting $\pi_K$ as the 2D projection operator onto the image plane,
we aim to minimize a reprojection error on the vertices relative to the ground truth $(K^*, V^*, t^*)$:
$\mathcal{L}_{\texttt{reproj}} = \frac{1}{|V|} \sum_n \big| \pi_K(\hat{\mathrm{V}}_n + \hat{t}_n) - \pi_{K^*}(\mathrm{V}^*_n+ t^*_n)\big|$.
For 50\% of the mini-batches, we randomly sample camera intrinsics with an horizontal field-of-view uniformly chosen between 5 and 170\textdegree{}.
For the remaining mini-batches, we use ground-truth camera intrinsics and introduce an additional objective function to minimize a human-centered vertices loss: $\mathcal{L}_{\texttt{mesh}} = \frac{1}{|V|} \sum_n \big| \hat{\mathrm{V}}_n - \mathrm{V}^*_n \big|$.
With the addition of these two deterministic losses, our total objective function can be expressed as:
\begin{equation}
\label{eq:loss}
    \mathcal{L} = \mathcal{L}_{\texttt{prob}} + \mathcal{L}_{\texttt{mesh}} + \mathcal{L}_{\texttt{reproj}}.
\end{equation}

\subsection{Implementation details}
\label{sec:impl}

\paragraph{Architecture.}
For our experiments, we use an architecture inspired by the single-shot framework of Multi-HMR~\cite{multi-hmr2024}, for its simplicity and state-of-the-art performance.
The input RGB image is encoded via a transformer-based architecture to produce 1024-dimensional image patch features $\mathcal{I}$, alongside  detection features of similar dimensions.
We train the entire network in an end-to-end manner to minimize the objective function~\eqref{eq:loss}, starting from DINOv2~\cite{oquab2023dinov2} weights initialization for the image encoder.
In our default settings, we use a ViT-Large encoder with an image resolution of $518\times518$ and a patch size of $14\times14$.
The MLPs outputting conditional distributions parameters take the conditioning variables as input and combine them through a sum in an hidden space (typically of dimension 256) after linear projection followed by a  rectilinear activation.
We use the Adam optimizer~\citep{kingma2015adam} with a learning rate of $5 \cdot 10^{-6}$ and train for 500k steps with a batch size of 16 images.
We consider all image patches with a detection score above 0.5 as detections in our experiments, after applying a non-maxima suppression of $3 \times 3$ patch window.

\paragraph{Training Data.}
We train models using only synthetic data in our experiments:
synthetic data has the advantage of mitigating personal privacy issues, it provides potentially perfect ground-truth annotations, and was shown to transfer well to real word applications in practice~\citep{black2023bedlam, multi-hmr2024,patel2021agora}.
Namely, we rely on the BEDLAM~\citep{black2023bedlam} dataset, which consists of 286k images depicting 951k persons.
To further increase the body shape diversity we additionally render a synthetic dataset of 8k scenes. It contains images depicting 7 persons on average, with 40 multi-view renderings per scene, leading to more than 300k images and 2M human instances rendered.
Additional details are provided in the supplementary material.

\paragraph{Computing Resources.}
\label{sec:computing}
Training our ViT-L model took about $51$ hours using one NVIDIA H100 GPU ($36h$ and $35.5$ hours resp. for ViT-B and ViT-S variants).
Inference time depends on the number of people visible in each image. For reference, we report average inference times on the 3DPW dataset in Table~\ref{tab:backbone}.

\section{Experiments}
\label{sec:experiments}

\begin{figure*}
\centering
\includegraphics[width=\textwidth]{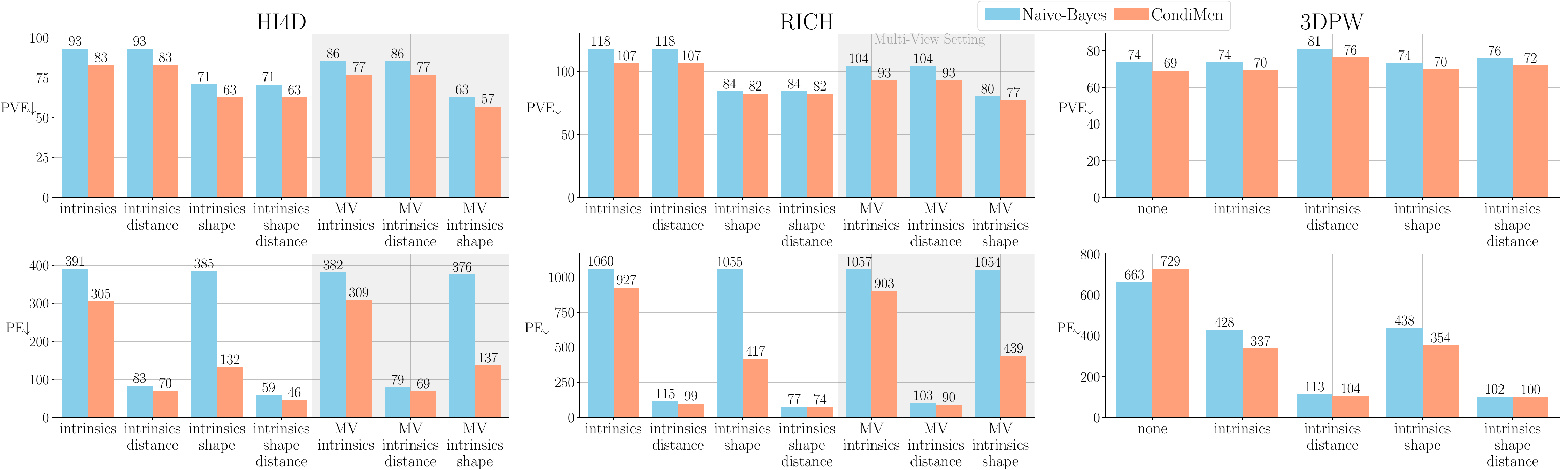}
\caption{\label{tab:mesh_recovery}%
\textbf{Impact of additional information}.
We report metrics for mesh reconstruction accuracy (Per Vertex Error, PVE) and 3D positioning (Positioning Error, PE) across different datasets.
Exploiting additional information such as camera intrinsics (\textit{intrinsics}), 
body shape parameters (\textit{shape}), distance to the camera (\textit{distance}), or multi-view observations (MV) can significantly reduce prediction errors.
By modeling relationships between variables, \emph{\ours} can consistently achieve lower errors than a \emph{Naive Bayes} baseline.%
}
\end{figure*}

To evaluate the effectiveness of our approach, we assess how \ours can leverage additional input information, and compare its performance against state-of-the-art methods.
We conduct experiments on single-view datasets (3DPW~\cite{3dpw} and MuPoTS~\cite{mupots}) as well as multi-view datasets (HI4D~\cite{rich}, Human3.6M~\cite{h36m} and RICH~\cite{rich}).
We report Mean Per Vertex Error in \emph{mm}, with or without Procrustes alignment (\emph{PVE} and \emph{PA-PVE}, resp.), along with mean absolute position error, (\emph{PE}) in \emph{mm}.
Following previous works~\cite{romp,bev,multi-hmr2024}, we also provide Mean Per Joint Error in \emph{mm} (\emph{PJE}) on 3DPW and Human3.6M, and Percentage of Correct Keypoints (\emph{PCK}) within $15cm$ %
on MuPoTS. For experiments with additional input information,
we convert SMPL annotations of 3DPW to the SMPL-X format using the code of \citet{choutas2020expose}.
Note that we do not perform these experiments on MuPoTS due to the absence of mesh annotations.

\paragraph{Graph connectivity.}
\label{sec:cond_prior}

Our Bayesian network architecture introduces conditional dependencies in the estimation of human attributes. For instance, pose prediction depends on body shape, as shown in the graphical model in  Fig.~\ref{fig:overview}.
To evaluate the effectiveness of this strategy, we compare our approach with a \emph{Naive-Bayes} baseline model, in which human attributes are predicted from detection features independently, without conditional dependencies, as illustrated in Fig.~\ref{fig:conditional_dependency_modeling}.
Such baseline is similar to Multi-HMR~\cite{multi-hmr2024}, except that it regresses parametric distributions instead of deterministic values, ensuring a more meaningful comparison with \emph{\ours}.
Results are summarized in Fig.~\ref{tab:mesh_recovery}, and additional results can be found in the supp.\ mat.
Our approach consistently outperforms the \emph{Naive-Bayes} baseline, %
especially when leveraging additional input information for certain variables.
This result underscores the importance of modeling dependencies between human attributes.

\paragraph{Additional input information.}
We report in Fig.~\ref{tab:mesh_recovery} performance gains achieved when 
incorporating additional input information, using the approach described in Sec.~\ref{sec:inference}.
As expected, providing distance to the camera in addition to the camera intrinsics (\emph{intrinsics}-\emph{distance}) significantly enhances %
mean absolute vertex position accuracy,
reducing the average PE error across datasets to 91mm for \emph{\ours} and 103mm for \emph{Naive Bayes}.
Exploiting body shape information (\emph{intrinsics-shape}) similarly boosts relative pose estimation, yielding up to a 25\% reduction in PVE error. %
It also brings significant improvement in term of absolute position error (-57\% on HI4D, -57\% on RICH  for \emph{\ours}), which suggests that the model is able to capture the relationship between visual appearance, body shape and distance to the camera, and to exploit this dependency to produce better predictions (see examples Fig.~\ref{fig:overview} and \ref{fig:3dpw_example1}). In contrast, the \emph{Naive Bayes} baseline which does not model attribute interdependencies shows only minor improvements in absolute position error %
(-1.6\% HI4D and -0.5\% on RICH).
On 3DPW, using additional shape information has %
marginal impact, as the model's initial shape estimates are already rather close to ground truth. %
However, using camera intrinsics significantly reduces localization error (by 53\% between \emph{none} and \emph{intrinsics}).
Finally, incorporating
external information regarding camera intrinsics, body shape and distance to the camera %
provides further improvements for both methods.

\paragraph{Multi-view.}
We also report in Fig.~\ref{tab:mesh_recovery} results obtained using a multi-view consistency prior, using the approach described in Sec.~\ref{sec:inference}.
We observe that the use of multiple views results in a smaller PVE error compared to the monocular case (see also Fig~\ref{fig:multiview_examples}), and that providing external input leads to further performance improvements, with observations similar to the ones made in the monocular case.

\newcommand{\multiviewmesh}[1]{\includegraphics[height=40pt,keepaspectratio]{#1}}
\begin{figure}
\centering
\footnotesize
\setlength{\tabcolsep}{1pt}
\begin{tabular}{cccc}
\includegraphics[width=0.13\textwidth,height=0.10\textwidth,keepaspectratio]{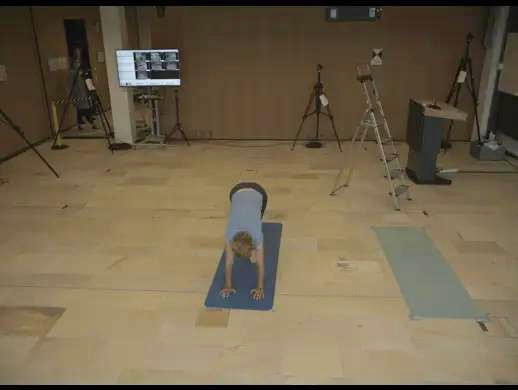}     &  \multiviewmesh{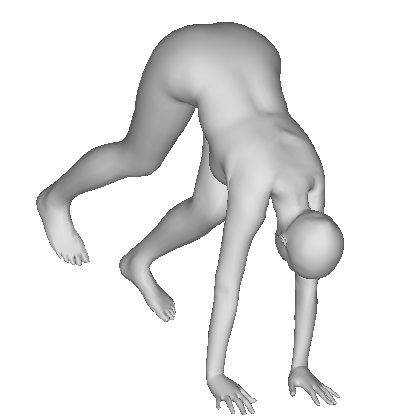} & \multiviewmesh{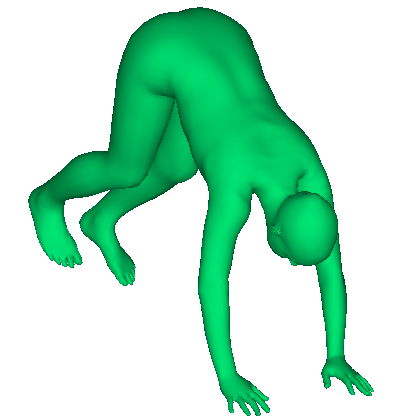} & \multiviewmesh{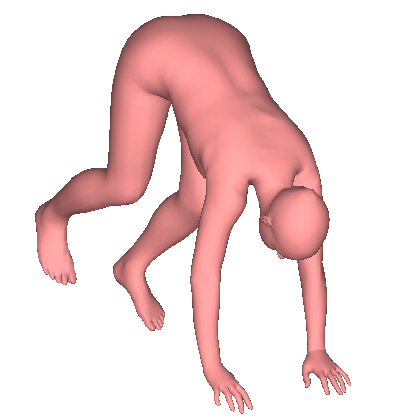}\\
(a) Input     & (b) Ground truth  & (c) Single-view & (d) Multi-view
\end{tabular}

\caption{\textbf{Improved prediction using multi-view prior.}}
\label{fig:multiview_examples}
\end{figure}

\begin{figure}[h]
	\scriptsize
	\centering
    \setlength{\tabcolsep}{1pt}
    \renewcommand{\arraystretch}{0.1}
    \newcommand{\mywildcolumn}[1]{\includegraphics[width=3cm,height=2.0cm,keepaspectratio]{figures/wild_results/#1/detection0.jpg} & \includegraphics[width=3cm,height=2.0cm,keepaspectratio]{figures/wild_results/#1/prediction0_.jpg} & \includegraphics[width=2cm,height=2.0cm,keepaspectratio]{figures/wild_results/#1/side_view.png}}
	\begin{tabular}{ccc}
		\mywildcolumn{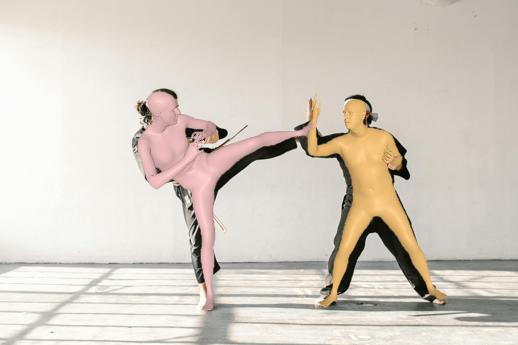} \\
		\mywildcolumn{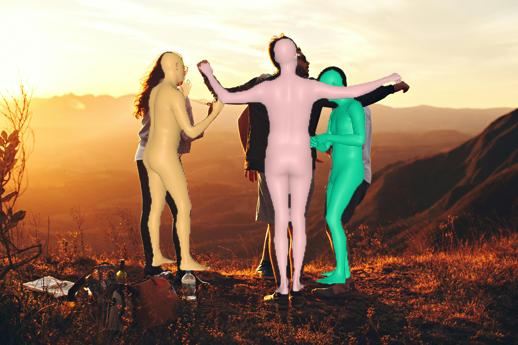} \\
		\mywildcolumn{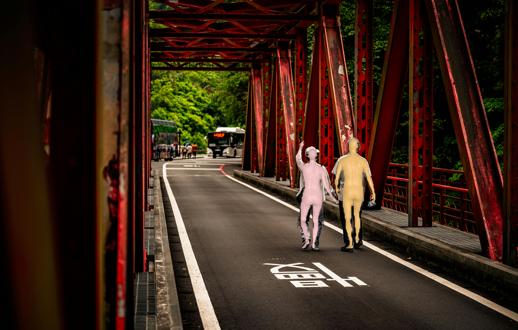} \\
		\mywildcolumn{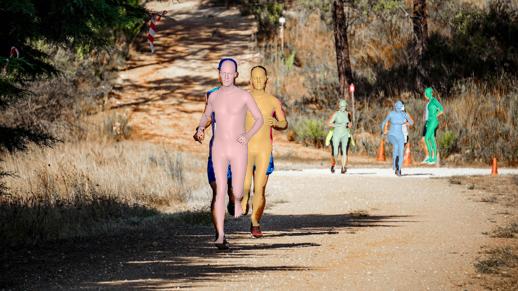} \\	
		\mywildcolumn{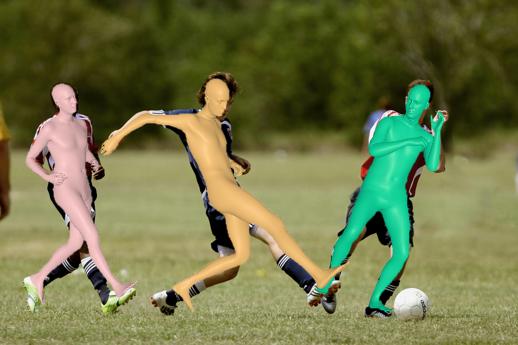} \\
		\mywildcolumn{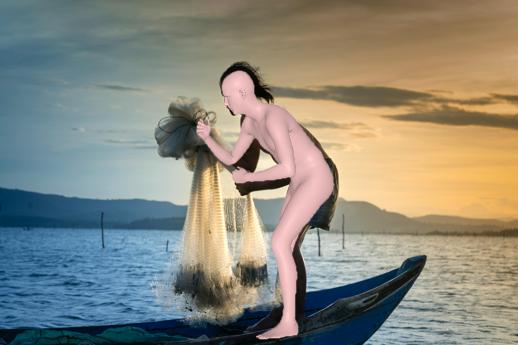}	\\
		\mywildcolumn{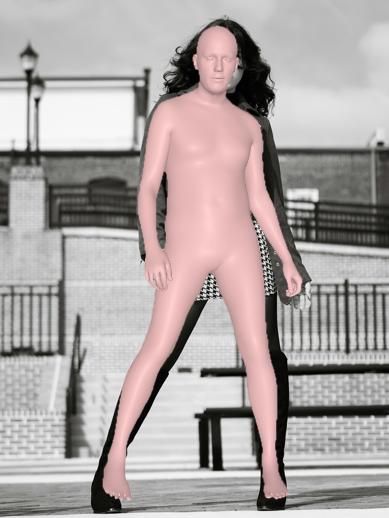} \\
        (a) Detection scores & (b) Prediction & (c) Side-view \\
	\end{tabular}

\caption{%
\textbf{Qualitative results} on free-to-use internet images.
Our method can produce plausible predictions even when no additional inputs (such as camera intrinsics or body shape) are available.
}
\label{fig:results_examples}
\end{figure}

\paragraph{State-of-the-art comparison.}
Finally, we compare the performance of our proposed method with recent approaches in both monocular and multi-view settings. %
 Table~\ref{tab:sota_comparison} presents the performance results as reported by the authors, except for Multi-HMR~\cite{multi-hmr2024}, which we retrained on our dataset using a ViT-Large backbone at $518{\times}518$ resolution for a fair comparison.
Following established practices~\cite{zhu2024muc}, we report results for Human3.6M, HI4D and RICH after fine-tuning on each corresponding train set (for $15k$ steps), and we use ground-truth camera intrinsics for evaluation~\cite{multi-hmr2024}.
We also report results of a strong multi-view baseline -- denoted \emph{Multi-HMR+avg} which averages (after a rigid registration) the shape and expression vectors, as well as the absolute bone orientations predicted by Multi-HMR across multiple views.
\ours achieves highly competitive results in both monocular and multi-view settings, and we provide qualitative examples in Fig.~\ref{fig:3dpw_example1} and \ref{fig:results_examples}.

\begin{table*}
\centering
\begin{minipage}{0.58\linewidth}
\centering
        \scriptsize
        \setlength{\tabcolsep}{3pt}
        \begin{tabular}{l|c|cccc|cc}
        \toprule
         \multirow{2}{*}{\bf{Method}} & \multirow{2}{*}{\bf{MV}} & \multicolumn{4}{c|}{\bf{Human3.6M} $\downarrow$}   & \multicolumn{2}{c}{\bf{HI4D} $\downarrow$} \\
        &  & PJE & PA-PJE & PVE & PA-PVE & PVE & PA-PVE \\
        \midrule
        ProHMR~\cite{kolotouros2021probabilistic} & & 65.1 & 43.7 & -- & -- \\ %
        ROMP~\cite{romp}   & &-- &-- & --& -- & 215.3 & -- \\ %
        BEV~\cite{bev}   & &-- &-- &-- &-- & 153.9 & -- \\  %
        HMR2.0~\cite{goel2023humans4d} & & \underline{50.0} & \textbf{32.4} & -- & -- & 141.2 & -- \\ %
        \citet{yu2022multiview} & & -- & 41.6 & -- & 46.4 & -- & -- \\ %
        SMPLer-X\cite{cai2023smplerx} & & -- & \underline{38.9} & -- & 42.8 & -- & -- \\ %
        MUC~\cite{zhu2024muc} & & -- & 44.3 & -- & 45.8 & -- & -- \\
        Multi-HMR~\cite{multi-hmr2024} && 50.8 & 40.0 & \underline{62.2} & \underline{43.3} & \underline{49.0} & \underline{36.2} \\ %
        \textbf{Ours} & & \textbf{49.0} & \underline{38.9} & \textbf{59.1} & \textbf{42.2} & \textbf{48.8} & \textbf{35.7} \\
        \midrule
        ProHMR~\cite{kolotouros2021probabilistic} & \checkmark & 62.2 & 34.5 & -- & -- & -- & -- \\ %
        \citet{yu2022multiview} & \checkmark & -- & 33.0 & -- & \underline{34.4} & -- & -- \\
        SMPLer-X\cite{cai2023smplerx} + avg & \checkmark & -- & 33.4 &  -- & 37.1 & -- & -- \\ %
        MUC~\cite{zhu2024muc} & \checkmark & -- & 31.9 & -- & 33.4 & -- & --  \\ %
        Calib-free PaFF~\cite{jia2023delving} & \checkmark & 44.8 & \underline{28.2} & -- & -- & -- & -- \\
        OUVR~\cite{xie2024visibility} & \checkmark & - & \textbf{27.1} & - & \textbf{28.9} & -- & -- \\
        Multi-HMR~\cite{multi-hmr2024} + avg & \checkmark & \underline{42.8} & 30.0 & 51.3 & 32.5 & \underline{51.1} & \underline{28.7} \\ %
        \textbf{Ours} & \checkmark & \textbf{41.1} & 28.9 & \textbf{49.0} & \underline{31.2} & \textbf{44.7} & \textbf{27.8} \\
        \bottomrule
        \end{tabular}
\subcaption{\textbf{Multi-view setting.} \label{tab:multiview_sota}}
\end{minipage}
\hfill
\begin{minipage}{0.4\linewidth}
\centering
\begin{minipage}{\textwidth}
\centering
\scriptsize
\setlength{\tabcolsep}{2pt}
\begin{tabular}{l|c|cc|c|c}
    \toprule
    \textbf{Method} & \multirow{2}{*}{\bf{MV}}& \multicolumn{2}{c|}{\textbf{Whole-body}$\downarrow$} & \textbf{Hands$\downarrow$} & \textbf{Face$\downarrow$} \\
    && PVE & PA-PVE & PA-PVE & PA-PVE \\
    \midrule
    MUC~\cite{zhu2024muc} && -- & 47.7 & 8.2 & 4.1 \\ %
    Multi-HMR~\cite{multi-hmr2024} & & \underline{69.4} & \underline{38.2} & \underline{7.4} & \underline{3.7} \\ %
    \textbf{Ours} && \textbf{65.4} & \textbf{36.5} & \textbf{7.3} & \textbf{3.6} \\
    \midrule
    MUC~\cite{zhu2024muc} & \checkmark & -- & 33.5 & \textbf{6.7} & \textbf{3.2} \\
    Multi-HMR~\cite{multi-hmr2024}+avg & \checkmark & \underline{61.7} & \underline{29.5} & 6.8 & 3.5 \\ %
    \textbf{Ours} &\checkmark & \textbf{57.8} & \textbf{27.9} &  \textbf{6.7} & \underline{3.4} \\ 
    \bottomrule
    \end{tabular}
\end{minipage}
\subcaption{\textbf{Body-part specific results on RICH.} \label{tab:part_specific_sota}}
\begin{minipage}{\textwidth}
\centering
\scriptsize
\setlength{\tabcolsep}{5.4pt}
\begin{tabular}{l|cc|cc}
    \toprule
    \multirow{2}{*}{\bf{Method}} & \multicolumn{2}{c|}{\bf{3DPW} $\downarrow$}   & \multicolumn{2}{c}{\textbf{MuPoTS PCK$\uparrow$}} \\
    & PJE & PA-PJE & Matched & All \\
    \midrule
    ProHMR~\cite{kolotouros2021probabilistic} & -- & 59.8 & -- & -- \\    
    ROMP~\cite{romp} & 76.7 & 47.3 & 69.9 & 72.2 \\ %
    BEV~\cite{bev} & 78.5 & 46.9 & 75.2 & 70.2 \\ %
    Multi-HMR~\cite{multi-hmr2024} & 70.6 & 47.5 & \underline{77.6} & \textbf{82.7} \\ %
    HMR2.0~\cite{goel2023humans4d} & 81.3 & 54.3 & -- & -- \\ %
    POCO~\cite{dwivedi2023poco} & \underline{69.7} & \textbf{42.8} & -- & -- \\ %
    \textbf{Ours}    & \textbf{69.5} & \underline{46.4} & \textbf{84.7} & \underline{74.0} \\ %
    \bottomrule
    \end{tabular}
\subcaption{\textbf{Monocular setting.}\label{tab:mono_sota}}
\end{minipage}
\end{minipage}
\caption{
\label{tab:sota_comparison}
\textbf{Comparison with state-of-the-art methods} in monocular and multi-view settings.
}
\end{table*}

\subsection{Additional ablations}
\label{sec:ablation}

\paragraph{Backbone.}
Tab.~\ref{tab:backbone} shows the results when experimenting with various image encoder backbones. As expected,
larger backbones yield better results: a ViT-Large model performs better than a ViT-Base which itself performs better than ViT-Small, likely due to the large scale of the training sets.
However, these improvements in prediction quality come  with additional computation costs both at training and inference time. Still, inference takes less than 50ms per image on average with a ViT-Large backbone, %
making it suitable for some real-time applications.

\paragraph{Training procedure.}
We also conduct ablation studies on various aspects of our training procedure, with results presented in Table~\ref{tab:training}.
First, we remove the mode-guiding objectives described in Sec.~\ref{sec:training} and observe 
a drop in
performance across all datasets.
We also train a variant of our model without the additional synthetic data we generated.
While this variant performs better on Human3.6M and HI4D, its performance is worse on RICH, 3DPW, and MuPoTS.
We posit this is due to the presence of less standard camera parameters or body shape attributes in these benchmarks, for which the diversity of additional training data proves beneficial. 

\begin{table}
\centering
\label{tabs:ablations}
\subfloat[\textbf{Backbone size.}\label{tab:backbone}]{
    \hspace{-0.2cm}
    \centering
    \scriptsize
    \setlength{\tabcolsep}{1.3pt}
    \resizebox{0.95\linewidth}{!}{
    \begin{tabular}{lcccccc}
    \toprule
    \bf{Backbone} & \textbf{Human3.6M$\downarrow$}& \textbf{HI4D$\downarrow$} & \textbf{RICH$\downarrow$} & \textbf{3DPW$\downarrow$} & \textbf{MuPoTS$\uparrow$} & \textbf{Inference$\downarrow$ (ms)}\\
    \midrule
    ViT-Small & 117.9 & 96.2 & 114.2 & 83.2 & 67.8 & \textbf{29} \\
    ViT-Base & \underline{99.7} & \underline{93.3} & \underline{108.2} & \underline{76.6} & \underline{69.7} & \underline{31} \\
    ViT-Large & \textbf{88.8} & \textbf{77.0} & \textbf{92.8} & \textbf{69.5} & \textbf{74.0} & 50\\
\bottomrule
    \end{tabular}
    }
}
\newline
\subfloat[\textbf{Training.}\label{tab:training}]{
    \hspace{-0.2cm}
    \centering
    \scriptsize
    \setlength{\tabcolsep}{1.3pt}
    \resizebox{0.95\linewidth}{!}{
    \scriptsize
    \begin{tabular}{lccccc}
    \toprule
    \textbf{Training} & \textbf{Human3.6M$\downarrow$}& \textbf{HI4D$\downarrow$} & \textbf{RICH$\downarrow$} & \textbf{3DPW$\downarrow$} & \textbf{MuPoTS$\uparrow$} \\
    \toprule
    \textbf{Ours} & 88.8 & 77.0 & \textbf{92.8} & \textbf{69.5} & \textbf{74.0} \\
    w/o mode guiding & 92.4 & 78.5 & 106.1 & 75.3 & 71.8 \\
    w/o additional synth. data & \textbf{75.9} & \textbf{73.8} & 96.4 & 74.4 & 71.5 \\
    \bottomrule
    \end{tabular}
    }
}
\caption{\textbf{Ablation study.} We report PVE metric in a multi-view setting on Human3.6M, HI4D and RICH. We report PJE for 3DPW and PCK-All for MuPoTS in a monocular setting. Inference durations %
are measured with an NVidia V100 GPU.
}
\end{table}

\paragraph{Uncertainty modeling} Empirically, we observe a correlation between the predicted likelihood $p(K, t, \theta, \beta, \gamma \vert \mathcal{I})$ and the prediction errors (see supp.\ mat.).
This suggests that the proposed model is able to estimate prediction uncertainty to some extent, an information that could be valuable for downstream applications.

\section{Discussion}
\label{sec:discussion}
We propose a novel approach for multi-person human mesh recovery, based on a Bayesian network. %
This method enables the seamless incorporation of additional input information -- such as
camera intrinsics, body shape, distance from the camera, or even multi-view acquisitions -- to improve the predictions.
We believe this has a significant practical 
value, as such data is readily available in many real-world applications.
Our approach achieves on par or better performances than existing state-of-the-art methods on standard benchmarks, while still
maintaining real-time capabilities.
The motivation behind our work was to exploit interdependencies between various attributes in the mesh recovery problem (\eg detection, camera parameters, and pose estimation).
To this end, we adopted a Bayesian network framework, which is conceptually simple and supports highly efficient inference.
Our findings confirms the effectiveness of this approach %
and suggest that 
exploring more sophisticated probabilistic modeling techniques could be a promising direction for future research.

{
    \small
    \bibliographystyle{ieeenat_fullname}
    \bibliography{biblio}

\begin{thebibliography}{85}
\providecommand{\natexlab}[1]{#1}
\providecommand{\url}[1]{\texttt{#1}}
\expandafter\ifx\csname urlstyle\endcsname\relax
  \providecommand{\doi}[1]{doi: #1}\else
  \providecommand{\doi}{doi: \begingroup \urlstyle{rm}\Url}\fi

\bibitem[ble()]{blender}
Blender.
\newblock \url{https://www.blender.org/}.

\bibitem[hum()]{humgen3d}
Humgen3d.
\newblock \url{https://www.humgen3d.com/}.

\bibitem[pol()]{polyhaven}
Poly haven.
\newblock \url{https://polyhaven.com/}.

\bibitem[Baradel et~al.(2024)Baradel, Armando, Galaaoui, Br{\'e}gier,
  Weinzaepfel, Rogez, and Lucas]{multi-hmr2024}
Fabien Baradel, Matthieu Armando, Salma Galaaoui, Romain Br{\'e}gier, Philippe
  Weinzaepfel, Gr{\'e}gory Rogez, and Thomas Lucas.
\newblock Multi-hmr: Multi-person whole-body human mesh recovery in a single
  shot.
\newblock In \emph{ECCV}, 2024.

\bibitem[Biggs et~al.(2020)Biggs, Novotny, Ehrhardt, Joo, Graham, and
  Vedaldi]{biggs20203d}
Benjamin Biggs, David Novotny, Sebastien Ehrhardt, Hanbyul Joo, Ben Graham, and
  Andrea Vedaldi.
\newblock 3d multi-bodies: Fitting sets of plausible 3d human models to
  ambiguous image data.
\newblock In \emph{NeurIPS}, 2020.

\bibitem[Bishop and Nasrabadi(2006)]{bishop2006pattern}
Christopher~M Bishop and Nasser~M Nasrabadi.
\newblock \emph{Pattern recognition and machine learning}.
\newblock Springer, 2006.

\bibitem[Black et~al.(2023)Black, Patel, Tesch, and Yang]{black2023bedlam}
Michael~J Black, Priyanka Patel, Joachim Tesch, and Jinlong Yang.
\newblock Bedlam: A synthetic dataset of bodies exhibiting detailed lifelike
  animated motion.
\newblock In \emph{CVPR}, 2023.

\bibitem[Bogo et~al.(2016)Bogo, Kanazawa, Lassner, Gehler, Romero, and
  Black]{bogo2016keep}
Federica Bogo, Angjoo Kanazawa, Christoph Lassner, Peter Gehler, Javier Romero,
  and Michael~J Black.
\newblock Keep it smpl: Automatic estimation of 3d human pose and shape from a
  single image.
\newblock In \emph{ECCV}, 2016.

\bibitem[Br{\'e}gier(2021)]{bregier2021deepregression}
Romain Br{\'e}gier.
\newblock Deep regression on manifolds: a {3D} rotation case study.
\newblock In \emph{3DV}, 2021.

\bibitem[Brown et~al.(2020)Brown, Mann, Ryder, Subbiah, Kaplan, Dhariwal,
  Neelakantan, Shyam, Sastry, Askell, et~al.]{brown2020language}
Tom Brown, Benjamin Mann, Nick Ryder, Melanie Subbiah, Jared~D Kaplan, Prafulla
  Dhariwal, Arvind Neelakantan, Pranav Shyam, Girish Sastry, Amanda Askell,
  et~al.
\newblock Language models are few-shot learners.
\newblock In \emph{NeurIPS}, 2020.

\bibitem[Cai et~al.(2023)Cai, Yin, Zeng, Wei, Sun, Wang, Pang, Mei, Zhang,
  Zhang, et~al.]{cai2023smplerx}
Zhongang Cai, Wanqi Yin, Ailing Zeng, Chen Wei, Qingping Sun, Yanjun Wang,
  Hui~En Pang, Haiyi Mei, Mingyuan Zhang, Lei Zhang, et~al.
\newblock Smpler-x: Scaling up expressive human pose and shape estimation.
\newblock In \emph{NeurIPS}, 2023.

\bibitem[Chang et~al.(2017)Chang, Dai, Funkhouser, Halber, Niessner, Savva,
  Song, Zeng, and Zhang]{matterport3d}
Angel Chang, Angela Dai, Thomas Funkhouser, Maciej Halber, Matthias Niessner,
  Manolis Savva, Shuran Song, Andy Zeng, and Yinda Zhang.
\newblock {Matterport3D}: Learning from {RGB-D} data in indoor environments.
\newblock In \emph{3DV}, 2017.

\bibitem[Chen et~al.(2016)Chen, Wang, Li, Su, Wang, Tu, Lischinski, Cohen-Or,
  and Chen]{chen2016synthesizing}
Wenzheng Chen, Huan Wang, Yangyan Li, Hao Su, Zhenhua Wang, Changhe Tu, Dani
  Lischinski, Daniel Cohen-Or, and Baoquan Chen.
\newblock Synthesizing training images for boosting human 3d pose estimation.
\newblock In \emph{3DV}, 2016.

\bibitem[Choudhury et~al.(2023)Choudhury, Kitani, and Jeni]{choudhury2023tempo}
Rohan Choudhury, Kris~M Kitani, and L{\'a}szl{\'o}~A Jeni.
\newblock Tempo: Efficient multi-view pose estimation, tracking, and
  forecasting.
\newblock In \emph{ICCV}, 2023.

\bibitem[Choutas et~al.(2020)Choutas, Pavlakos, Bolkart, Tzionas, and
  Black]{choutas2020expose}
Vasileios Choutas, Georgios Pavlakos, Timo Bolkart, Dimitrios Tzionas, and
  Michael~J Black.
\newblock Monocular expressive body regression through body-driven attention.
\newblock In \emph{ECCV}, 2020.

\bibitem[Davoodnia et~al.(2024)Davoodnia, Ghorbani, Carbonneau, Messier, and
  Etemad]{davoodnia2025upose3d}
Vandad Davoodnia, Saeed Ghorbani, Marc-Andr{\'e} Carbonneau, Alexandre Messier,
  and Ali Etemad.
\newblock Upose3d: Uncertainty-aware 3d human pose estimation with cross-view
  and temporal cues.
\newblock In \emph{ECCV}, 2024.

\bibitem[Dosovitskiy et~al.(2021)Dosovitskiy, Beyer, Kolesnikov, Weissenborn,
  Zhai, Unterthiner, Dehghani, Minderer, Heigold, Gelly,
  et~al.]{dosovitskiy2020image}
Alexey Dosovitskiy, Lucas Beyer, Alexander Kolesnikov, Dirk Weissenborn,
  Xiaohua Zhai, Thomas Unterthiner, Mostafa Dehghani, Matthias Minderer, Georg
  Heigold, Sylvain Gelly, et~al.
\newblock An image is worth 16x16 words: Transformers for image recognition at
  scale.
\newblock In \emph{ICLR}, 2021.

\bibitem[Dwivedi et~al.(2024{\natexlab{a}})Dwivedi, Schmid, Yi, Black, and
  Tzionas]{dwivedi2023poco}
Sai~Kumar Dwivedi, Cordelia Schmid, Hongwei Yi, Michael~J Black, and Dimitrios
  Tzionas.
\newblock Poco: 3d pose and shape estimation with confidence.
\newblock In \emph{3DV}, 2024{\natexlab{a}}.

\bibitem[Dwivedi et~al.(2024{\natexlab{b}})Dwivedi, Sun, Patel, Feng, and
  Black]{dwivedi2024tokenhmr}
Sai~Kumar Dwivedi, Yu Sun, Priyanka Patel, Yao Feng, and Michael~J Black.
\newblock Tokenhmr: Advancing human mesh recovery with a tokenized pose
  representation.
\newblock In \emph{CVPR}, 2024{\natexlab{b}}.

\bibitem[Esser et~al.(2021)Esser, Rombach, and Ommer]{esser2021taming}
Patrick Esser, Robin Rombach, and Bjorn Ommer.
\newblock Taming transformers for high-resolution image synthesis.
\newblock In \emph{CVPR}, 2021.

\bibitem[Feng et~al.(2021)Feng, Choutas, Bolkart, Tzionas, and Black]{pixie}
Yao Feng, Vasileios Choutas, Timo Bolkart, Dimitrios Tzionas, and Michael~J
  Black.
\newblock Collaborative regression of expressive bodies using moderation.
\newblock In \emph{3DV}, 2021.

\bibitem[Fiche et~al.(2024)Fiche, Leglaive, Alameda-Pineda, Agudo, and
  Moreno-Noguer]{fiche2023vq}
Gu{\'e}nol{\'e} Fiche, Simon Leglaive, Xavier Alameda-Pineda, Antonio Agudo,
  and Francesc Moreno-Noguer.
\newblock Vq-hps: Human pose and shape estimation in a vector-quantized latent
  space.
\newblock \emph{ECCV}, 2024.

\bibitem[Goel et~al.(2023)Goel, Pavlakos, Rajasegaran, Kanazawa, and
  Malik]{goel2023humans4d}
Shubham Goel, Georgios Pavlakos, Jathushan Rajasegaran, Angjoo Kanazawa, and
  Jitendra Malik.
\newblock Humans in 4d: Reconstructing and tracking humans with transformers.
\newblock In \emph{ICCV}, 2023.

\bibitem[Hewitt et~al.(2024)Hewitt, Saleh, Aliakbarian, Petikam, Rezaeifar,
  Florentin, Hosenie, Cashman, Valentin, Cosker, et~al.]{hewitt2024look}
Charlie Hewitt, Fatemeh Saleh, Sadegh Aliakbarian, Lohit Petikam, Shideh
  Rezaeifar, Louis Florentin, Zafiirah Hosenie, Thomas~J Cashman, Julien
  Valentin, Darren Cosker, et~al.
\newblock Look ma, no markers: holistic performance capture without the hassle.
\newblock In \emph{SIGGRAPH Asia}, 2024.

\bibitem[Huang et~al.(2024)Huang, Li, Xu, Pan, Wang, and Lee]{huang2024closely}
Buzhen Huang, Chen Li, Chongyang Xu, Liang Pan, Yangang Wang, and Gim~Hee Lee.
\newblock Closely interactive human reconstruction with proxemics and
  physics-guided adaption.
\newblock In \emph{CVPR}, 2024.

\bibitem[Huang et~al.(2022)Huang, Yi, H{\"o}schle, Safroshkin, Alexiadis,
  Polikovsky, Scharstein, and Black]{rich}
Chun-Hao~P. Huang, Hongwei Yi, Markus H{\"o}schle, Matvey Safroshkin,
  Tsvetelina Alexiadis, Senya Polikovsky, Daniel Scharstein, and Michael~J.
  Black.
\newblock Capturing and inferring dense full-body human-scene contact.
\newblock In \emph{CVPR}, 2022.

\bibitem[Ionescu et~al.(2014)Ionescu, Papava, Olaru, and Sminchisescu]{h36m}
Catalin Ionescu, Dragos Papava, Vlad Olaru, and Cristian Sminchisescu.
\newblock Human3.6m: Large scale datasets and predictive methods for 3d human
  sensing in natural environments.
\newblock \emph{IEEE Trans. PAMI}, 2014.

\bibitem[Iskakov et~al.(2019)Iskakov, Burkov, Lempitsky, and
  Malkov]{iskakov2019learnable}
Karim Iskakov, Egor Burkov, Victor Lempitsky, and Yury Malkov.
\newblock Learnable triangulation of human pose.
\newblock In \emph{ICCV}, 2019.

\bibitem[Jahangiri and Yuille(2017)]{jahangiri2017generating}
Ehsan Jahangiri and Alan~L Yuille.
\newblock Generating multiple diverse hypotheses for human 3d pose consistent
  with 2d joint detections.
\newblock In \emph{ICCVW}, 2017.

\bibitem[Jia et~al.(2023)Jia, Zhang, An, and Liu]{jia2023delving}
Kai Jia, Hongwen Zhang, Liang An, and Yebin Liu.
\newblock Delving deep into pixel alignment feature for accurate multi-view
  human mesh recovery.
\newblock In \emph{AAAI}, 2023.

\bibitem[Joo et~al.(2021)Joo, Neverova, and Vedaldi]{joo2021exemplar}
Hanbyul Joo, Natalia Neverova, and Andrea Vedaldi.
\newblock Exemplar fine-tuning for 3d human model fitting towards in-the-wild
  3d human pose estimation.
\newblock In \emph{3DV}, 2021.

\bibitem[Kanazawa et~al.(2018)Kanazawa, Black, Jacobs, and
  Malik]{kanazawa2018hmr}
Angjoo Kanazawa, Michael~J. Black, David~W. Jacobs, and Jitendra Malik.
\newblock End-to-end recovery of human shape and pose.
\newblock In \emph{CVPR}, 2018.

\bibitem[Kingma and Ba(2015)]{kingma2015adam}
Diederik~P. Kingma and Jimmy Ba.
\newblock Adam: A method for stochastic optimization.
\newblock In \emph{{ICLR}}, 2015.

\bibitem[Kolotouros et~al.(2019)Kolotouros, Pavlakos, Black, and
  Daniilidis]{kolotouros2019spin}
Nikos Kolotouros, Georgios Pavlakos, Michael~J Black, and Kostas Daniilidis.
\newblock Learning to reconstruct 3d human pose and shape via model-fitting in
  the loop.
\newblock In \emph{ICCV}, 2019.

\bibitem[Kolotouros et~al.(2021)Kolotouros, Pavlakos, Jayaraman, and
  Daniilidis]{kolotouros2021probabilistic}
Nikos Kolotouros, Georgios Pavlakos, Dinesh Jayaraman, and Kostas Daniilidis.
\newblock Probabilistic modeling for human mesh recovery.
\newblock In \emph{ICCV}, 2021.

\bibitem[Li and Lee(2019)]{li2019generating}
Chen Li and Gim~Hee Lee.
\newblock Generating multiple hypotheses for 3d human pose estimation with
  mixture density network.
\newblock In \emph{CVPR}, 2019.

\bibitem[Li et~al.(2021{\natexlab{a}})Li, Bian, Zeng, Wang, Pang, Liu, and
  Lu]{li2021human}
Jiefeng Li, Siyuan Bian, Ailing Zeng, Can Wang, Bo Pang, Wentao Liu, and Cewu
  Lu.
\newblock Human pose regression with residual log-likelihood estimation.
\newblock In \emph{ICCV}, 2021{\natexlab{a}}.

\bibitem[Li et~al.(2021{\natexlab{b}})Li, Oskarsson, and Heyden]{li20213d}
Zhongguo Li, Magnus Oskarsson, and Anders Heyden.
\newblock 3d human pose and shape estimation through collaborative learning and
  multi-view model-fitting.
\newblock In \emph{WACV}, 2021{\natexlab{b}}.

\bibitem[Li et~al.(2022)Li, Liu, Zhang, Xu, and Yan]{li2022cliff}
Zhihao Li, Jianzhuang Liu, Zhensong Zhang, Songcen Xu, and Youliang Yan.
\newblock Cliff: Carrying location information in full frames into human pose
  and shape estimation.
\newblock In \emph{ECCV}, 2022.

\bibitem[Lin et~al.(2023)Lin, Zeng, Wang, Zhang, and Li]{osx}
Jing Lin, Ailing Zeng, Haoqian Wang, Lei Zhang, and Yu Li.
\newblock One-stage 3d whole-body mesh recovery with component aware
  transformer.
\newblock In \emph{CVPR}, 2023.

\bibitem[Loper et~al.(2015)Loper, Mahmood, Romero, Pons-Moll, and
  Black]{SMPL:2015}
Matthew Loper, Naureen Mahmood, Javier Romero, Gerard Pons-Moll, and Michael~J.
  Black.
\newblock {SMPL}: A skinned multi-person linear model.
\newblock \emph{ACM Trans. Graphics}, 2015.

\bibitem[Mehta et~al.(2018)Mehta, Sotnychenko, Mueller, Xu, Sridhar, Pons-Moll,
  and Theobalt]{mupots}
Dushyant Mehta, Oleksandr Sotnychenko, Franziska Mueller, Weipeng Xu, Srinath
  Sridhar, Gerard Pons-Moll, and Christian Theobalt.
\newblock Single-shot multi-person 3d pose estimation from monocular rgb.
\newblock In \emph{3DV}, 2018.

\bibitem[Moon et~al.(2022)Moon, Choi, and Lee]{hand4whole}
Gyeongsik Moon, Hongsuk Choi, and Kyoung~Mu Lee.
\newblock Accurate 3d hand pose estimation for whole-body 3d human mesh
  estimation.
\newblock In \emph{CVPRW}, 2022.

\bibitem[Müller et~al.(2024)Müller, Ye, Pavlakos, Black, and
  Kanazawa]{muller2024buddi}
Lea Müller, Vickie Ye, Georgios Pavlakos, Michael Black, and Angjoo Kanazawa.
\newblock Generative {Proxemics}: {A} {Prior} for {3D} {Social} {Interaction}
  from {Images}.
\newblock In \emph{{CVPR}}, 2024.

\bibitem[Oquab et~al.(2024)Oquab, Darcet, Moutakanni, Vo, Szafraniec, Khalidov,
  Fernandez, Haziza, Massa, El-Nouby, et~al.]{oquab2023dinov2}
Maxime Oquab, Timoth{\'e}e Darcet, Th{\'e}o Moutakanni, Huy Vo, Marc
  Szafraniec, Vasil Khalidov, Pierre Fernandez, Daniel Haziza, Francisco Massa,
  Alaaeldin El-Nouby, et~al.
\newblock Dinov2: Learning robust visual features without supervision.
\newblock \emph{TMLR}, 2024.

\bibitem[Patel et~al.(2021)Patel, Huang, Tesch, Hoffmann, Tripathi, and
  Black]{patel2021agora}
Priyanka Patel, Chun-Hao~P Huang, Joachim Tesch, David~T Hoffmann, Shashank
  Tripathi, and Michael~J Black.
\newblock Agora: Avatars in geography optimized for regression analysis.
\newblock In \emph{CVPR}, 2021.

\bibitem[Pavlakos et~al.(2019{\natexlab{a}})Pavlakos, Choutas, Ghorbani,
  Bolkart, Osman, Tzionas, and Black]{ehf}
Georgios Pavlakos, Vasileios Choutas, Nima Ghorbani, Timo Bolkart, Ahmed~AA
  Osman, Dimitrios Tzionas, and Michael~J Black.
\newblock Expressive body capture: 3d hands, face, and body from a single
  image.
\newblock In \emph{CVPR}, 2019{\natexlab{a}}.

\bibitem[Pavlakos et~al.(2019{\natexlab{b}})Pavlakos, Choutas, Ghorbani,
  Bolkart, Osman, Tzionas, and Black]{pavlakos2019expressive}
Georgios Pavlakos, Vasileios Choutas, Nima Ghorbani, Timo Bolkart, Ahmed~AA
  Osman, Dimitrios Tzionas, and Michael~J Black.
\newblock Expressive body capture: 3d hands, face, and body from a single
  image.
\newblock In \emph{CVPR}, 2019{\natexlab{b}}.

\bibitem[Pavlakos et~al.(2022)Pavlakos, Malik, and Kanazawa]{pavlakos2022human}
Georgios Pavlakos, Jitendra Malik, and Angjoo Kanazawa.
\newblock Human mesh recovery from multiple shots.
\newblock In \emph{CVPR}, 2022.

\bibitem[Qiu et~al.(2019)Qiu, Wang, Wang, Wang, and Zeng]{qiu2019cross}
Haibo Qiu, Chunyu Wang, Jingdong Wang, Naiyan Wang, and Wenjun Zeng.
\newblock Cross view fusion for 3d human pose estimation.
\newblock In \emph{ICCV}, 2019.

\bibitem[Qiu et~al.(2023)Qiu, Yang, Wang, Feng, Han, Ding, Xu, Fu, and
  Wang]{psvt}
Zhongwei Qiu, Qiansheng Yang, Jian Wang, Haocheng Feng, Junyu Han, Errui Ding,
  Chang Xu, Dongmei Fu, and Jingdong Wang.
\newblock Psvt: End-to-end multi-person 3d pose and shape estimation with
  progressive video transformers.
\newblock In \emph{CVPR}, 2023.

\bibitem[Rogez et~al.(2017)Rogez, Weinzaepfel, and Schmid]{rogez2017lcr}
Gregory Rogez, Philippe Weinzaepfel, and Cordelia Schmid.
\newblock Lcr-net: Localization-classification-regression for human pose.
\newblock In \emph{CVPR}, 2017.

\bibitem[Rong et~al.(2021)Rong, Shiratori, and Joo]{rong2021frankmocap}
Yu Rong, Takaaki Shiratori, and Hanbyul Joo.
\newblock Frankmocap: A monocular 3d whole-body pose estimation system via
  regression and integration.
\newblock In \emph{ICCV}, 2021.

\bibitem[S{\'a}r{\'a}ndi and Pons-Moll(2024)]{sarandi2024neural}
Istv{\'a}n S{\'a}r{\'a}ndi and Gerard Pons-Moll.
\newblock Neural localizer fields for continuous 3d human pose and shape
  estimation.
\newblock \emph{NeurIPS}, 2024.

\bibitem[Savva et~al.(2019)Savva, Kadian, Maksymets, Zhao, Wijmans, Jain,
  Straub, Liu, Koltun, Malik, et~al.]{habitat}
Manolis Savva, Abhishek Kadian, Oleksandr Maksymets, Yili Zhao, Erik Wijmans,
  Bhavana Jain, Julian Straub, Jia Liu, Vladlen Koltun, Jitendra Malik, et~al.
\newblock Habitat: A platform for embodied ai research.
\newblock In \emph{ICCV}, 2019.

\bibitem[Sengupta et~al.(2021{\natexlab{a}})Sengupta, Budvytis, and
  Cipolla]{sengupta2021hierarchical}
Akash Sengupta, Ignas Budvytis, and Roberto Cipolla.
\newblock Hierarchical kinematic probability distributions for 3d human shape
  and pose estimation from images in the wild.
\newblock In \emph{ICCV}, 2021{\natexlab{a}}.

\bibitem[Sengupta et~al.(2021{\natexlab{b}})Sengupta, Budvytis, and
  Cipolla]{sengupta2021probabilistic}
Akash Sengupta, Ignas Budvytis, and Roberto Cipolla.
\newblock Probabilistic 3d human shape and pose estimation from multiple
  unconstrained images in the wild.
\newblock In \emph{CVPR}, 2021{\natexlab{b}}.

\bibitem[Sengupta et~al.(2023)Sengupta, Budvytis, and
  Cipolla]{sengupta2023humaniflow}
Akash Sengupta, Ignas Budvytis, and Roberto Cipolla.
\newblock Humaniflow: Ancestor-conditioned normalising flows on so (3)
  manifolds for human pose and shape distribution estimation.
\newblock In \emph{CVPR}, 2023.

\bibitem[Shin et~al.(2024)Shin, Kim, Halilaj, and Black]{shin2024wham}
Soyong Shin, Juyong Kim, Eni Halilaj, and Michael~J Black.
\newblock Wham: Reconstructing world-grounded humans with accurate 3d motion.
\newblock In \emph{CVPR}, 2024.

\bibitem[Sidenbladh et~al.(2000)Sidenbladh, Black, and
  Fleet]{sidenbladh2000stochastic}
Hedvig Sidenbladh, Michael~J Black, and David~J Fleet.
\newblock Stochastic tracking of 3d human figures using 2d image motion.
\newblock In \emph{ECCV}, 2000.

\bibitem[Sminchisescu and Triggs(2001)]{sminchisescu2001covariance}
Cristian Sminchisescu and Bill Triggs.
\newblock Covariance scaled sampling for monocular 3d body tracking.
\newblock In \emph{CVPR}, 2001.

\bibitem[Sun et~al.(2024)Sun, Wang, Zeng, Yin, Wei, Wang, Mei, Leung, Liu,
  Yang, et~al.]{sun2024aios}
Qingping Sun, Yanjun Wang, Ailing Zeng, Wanqi Yin, Chen Wei, Wenjia Wang, Haiyi
  Mei, Chi-Sing Leung, Ziwei Liu, Lei Yang, et~al.
\newblock Aios: All-in-one-stage expressive human pose and shape estimation.
\newblock In \emph{CVPR}, 2024.

\bibitem[Sun et~al.(2021)Sun, Bao, Liu, Fu, Black, and Mei]{romp}
Yu Sun, Qian Bao, Wu Liu, Yili Fu, Michael~J Black, and Tao Mei.
\newblock Monocular, one-stage, regression of multiple 3d people.
\newblock In \emph{ICCV}, 2021.

\bibitem[Sun et~al.(2022)Sun, Liu, Bao, Fu, Mei, and Black]{bev}
Yu Sun, Wu Liu, Qian Bao, Yili Fu, Tao Mei, and Michael~J Black.
\newblock Putting people in their place: Monocular regression of 3d people in
  depth.
\newblock In \emph{CVPR}, 2022.

\bibitem[Tu et~al.(2020)Tu, Wang, and Zeng]{tu2020voxelpose}
Hanyue Tu, Chunyu Wang, and Wenjun Zeng.
\newblock Voxelpose: Towards multi-camera 3d human pose estimation in wild
  environment.
\newblock In \emph{ECCV}, 2020.

\bibitem[Van~den Oord et~al.(2016)Van~den Oord, Kalchbrenner, Espeholt,
  Vinyals, Graves, et~al.]{van2016conditional}
Aaron Van~den Oord, Nal Kalchbrenner, Lasse Espeholt, Oriol Vinyals, Alex
  Graves, et~al.
\newblock Conditional image generation with pixelcnn decoders.
\newblock In \emph{NeurIPS}, 2016.

\bibitem[von Marcard et~al.(2018)von Marcard, Henschel, Black, Rosenhahn, and
  Pons-Moll]{3dpw}
Timo von Marcard, Roberto Henschel, Michael~J. Black, Bodo Rosenhahn, and
  Gerard Pons-Moll.
\newblock Recovering accurate 3d human pose in the wild using imus and a moving
  camera.
\newblock In \emph{ECCV}, 2018.

\bibitem[Wang et~al.(2024)Wang, Wang, Liu, and Daniilidis]{wang2025tram}
Yufu Wang, Ziyun Wang, Lingjie Liu, and Kostas Daniilidis.
\newblock Tram: Global trajectory and motion of 3d humans from in-the-wild
  videos.
\newblock In \emph{ECCV}, 2024.

\bibitem[Wehrbein et~al.(2021)Wehrbein, Rudolph, Rosenhahn, and
  Wandt]{wehrbein2021probabilistic}
Tom Wehrbein, Marco Rudolph, Bodo Rosenhahn, and Bastian Wandt.
\newblock Probabilistic monocular 3d human pose estimation with normalizing
  flows.
\newblock In \emph{ICCV}, 2021.

\bibitem[Weinzaepfel et~al.(2020)Weinzaepfel, Br\'egier, Combaluzier, Leroy,
  and Rogez]{dope}
Philippe Weinzaepfel, Romain Br\'egier, Hadrien Combaluzier, Vincent Leroy, and
  Gr\'egory Rogez.
\newblock Dope: Distillation of part experts for whole-body 3d pose estimation
  in the wild.
\newblock In \emph{ECCV}, 2020.

\bibitem[Xia et~al.(2018)Xia, R.~Zamir, He, Sax, Malik, and
  Savarese]{gibsonenv}
Fei Xia, Amir R.~Zamir, Zhi-Yang He, Alexander Sax, Jitendra Malik, and Silvio
  Savarese.
\newblock Gibson {Env}: real-world perception for embodied agents.
\newblock In \emph{CVPR}, 2018.

\bibitem[Xie et~al.(2024)Xie, He, Zou, Wu, Liu, Zhao, Wang, Lin, and
  Lin]{xie2024visibility}
Zhenyu Xie, Huanyu He, Gui Zou, Jie Wu, Guoliang Liu, Jun Zhao, Yingxue Wang,
  Hui Lin, and Weiyao Lin.
\newblock Visibility-guided human body reconstruction from uncalibrated
  multi-view cameras.
\newblock In \emph{ICMR}, 2024.

\bibitem[Xu et~al.(2020)Xu, Bazavan, Zanfir, Freeman, Sukthankar, and
  Sminchisescu]{xu2020ghum}
Hongyi Xu, Eduard~Gabriel Bazavan, Andrei Zanfir, William~T Freeman, Rahul
  Sukthankar, and Cristian Sminchisescu.
\newblock Ghum \& ghuml: Generative 3d human shape and articulated pose models.
\newblock In \emph{CVPR}, 2020.

\bibitem[Ye et~al.(2022)Ye, Zhu, Wang, Wu, and Wang]{ye2022faster}
Hang Ye, Wentao Zhu, Chunyu Wang, Rujie Wu, and Yizhou Wang.
\newblock Faster voxelpose: Real-time 3d human pose estimation by orthographic
  projection.
\newblock In \emph{ECCV}, 2022.

\bibitem[Yershova et~al.(2010)Yershova, Jain, Lavalle, and
  Mitchell]{yershova2010so3grid}
Anna Yershova, Swati Jain, Steven~M Lavalle, and Julie~C Mitchell.
\newblock Generating uniform incremental grids on so (3) using the hopf
  fibration.
\newblock \emph{IJRR}, 2010.

\bibitem[Yin et~al.(2023)Yin, Guo, Kaufmann, Zarate, Song, and Hilliges]{hi4d}
Yifei Yin, Chen Guo, Manuel Kaufmann, Juan Zarate, Jie Song, and Otmar
  Hilliges.
\newblock Hi4d: 4d instance segmentation of close human interaction.
\newblock In \emph{CVPR}, 2023.

\bibitem[Yu et~al.(2022)Yu, Zhang, Xu, Tang, Tran, Keskin, and
  Park]{yu2022multiview}
Zhixuan Yu, Linguang Zhang, Yuanlu Xu, Chengcheng Tang, Luan Tran, Cem Keskin,
  and Hyun~Soo Park.
\newblock Multiview human body reconstruction from uncalibrated cameras.
\newblock In \emph{NeurIPS}, 2022.

\bibitem[Yuan et~al.(2023)Yuan, Song, Iqbal, Vahdat, and
  Kautz]{yuan2023physdiff}
Ye Yuan, Jiaming Song, Umar Iqbal, Arash Vahdat, and Jan Kautz.
\newblock Physdiff: Physics-guided human motion diffusion model.
\newblock In \emph{ICCV}, 2023.

\bibitem[Zhang et~al.(2021)Zhang, Tian, Zhou, Ouyang, Liu, Wang, and
  Sun]{zhang2021pymaf}
Hongwen Zhang, Yating Tian, Xinchi Zhou, Wanli Ouyang, Yebin Liu, Limin Wang,
  and Zhenan Sun.
\newblock Pymaf: 3d human pose and shape regression with pyramidal mesh
  alignment feedback loop.
\newblock In \emph{ICCV}, 2021.

\bibitem[Zhang et~al.(2023{\natexlab{a}})Zhang, Tian, Zhang, Li, An, Sun, and
  Liu]{pymafx2023}
Hongwen Zhang, Yating Tian, Yuxiang Zhang, Mengcheng Li, Liang An, Zhenan Sun,
  and Yebin Liu.
\newblock Pymaf-x: Towards well-aligned full-body model regression from
  monocular images.
\newblock \emph{IEEE Trans. PAMI}, 2023{\natexlab{a}}.

\bibitem[Zhang et~al.(2022)Zhang, Chen, and Tu]{zhang2022uncertainty}
Jinlu Zhang, Yujin Chen, and Zhigang Tu.
\newblock Uncertainty-aware 3d human pose estimation from monocular video.
\newblock In \emph{ACMMM}, 2022.

\bibitem[Zhang et~al.(2023{\natexlab{b}})Zhang, Ma, Zhang, Aliakbarian, Cosker,
  and Tang]{zhang2023probabilistic}
Siwei Zhang, Qianli Ma, Yan Zhang, Sadegh Aliakbarian, Darren Cosker, and Siyu
  Tang.
\newblock Probabilistic human mesh recovery in 3d scenes from egocentric views.
\newblock In \emph{ICCV}, 2023{\natexlab{b}}.

\bibitem[Zhou et~al.(2019)Zhou, Wang, and Kr{\"a}henb{\"u}hl]{centernet}
Xingyi Zhou, Dequan Wang, and Philipp Kr{\"a}henb{\"u}hl.
\newblock Objects as points.
\newblock In \emph{arXiv preprint arXiv:1904.07850}, 2019.

\bibitem[Zhou et~al.(2021)Zhou, Habermann, Habibie, Tewari, Theobalt, and
  Xu]{zhou2021monocular}
Yuxiao Zhou, Marc Habermann, Ikhsanul Habibie, Ayush Tewari, Christian
  Theobalt, and Feng Xu.
\newblock Monocular real-time full body capture with inter-part correlations.
\newblock In \emph{CVPR}, 2021.

\bibitem[Zhu et~al.(2024)Zhu, Wang, Xu, Zhuang, Wang, Wang, Zhang, and
  Wang]{zhu2024muc}
Yitao Zhu, Sheng Wang, Mengjie Xu, Zixu Zhuang, Zhixin Wang, Kaidong Wang, Han
  Zhang, and Qian Wang.
\newblock Muc: Mixture of uncalibrated cameras for robust 3d human body
  reconstruction.
\newblock \emph{arXiv preprint arXiv:2403.05055}, 2024.

\end{thebibliography}
}

\clearpage
\appendix
\begin{table*}
    \centering
    \Large
    \bf
    Supplementary material
\end{table*}

This document provides additional material regarding \emph{\Ours: Conditional Multi-Person Mesh Recovery}.
In Sec.~\ref{sec:counterfactual_xps}, \ref{sec:uncertainty} and \ref{sec:failure_cases} we report results of additional experiments aiming at better characterizing properties of \Ours.
Sec.~\ref{sec:connectivity_ablation} contains experimental results considering different Bayesian network connectivity, complementing results presented in the main paper.
Lastly, in Sec.~\ref{sec:implementation_details} we describe some implementation details used in our experiments.

\section{Attributes dependency modeling}
\label{sec:counterfactual_xps}
\begin{figure*}[!h]
\centering
\newcommand{\mupotstrimmedgraphics}[1]{\includegraphics[height=2cm,trim={2cm 0 4cm 0},clip]{#1}}
\newcommand{\threedpwtrimmedgraphics}[1]{\includegraphics[height=2cm,trim={2.0cm 5mm 5.0cm 0},clip]{#1}}
\newcommand{\threedpwtrimmedgraphicsbis}[1]{\includegraphics[height=2cm,trim={4.8cm 5mm 2.5cm 0},clip]{#1}}
\newcommand{\hifourdtrimmedgraphics}[1]{\includegraphics[height=2.4cm,trim={2.5mm 0 4.0mm 0},clip]{#1}}

\renewcommand{\arraystretch}{1.5}
\setlength{\tabcolsep}{1pt}

\begin{subfigure}{\linewidth}
\centering
\begin{tabular}{cccccc}
\mupotstrimmedgraphics{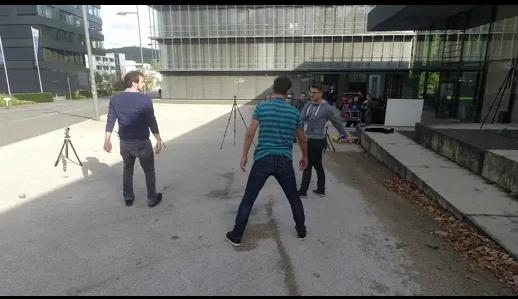} &
\mupotstrimmedgraphics{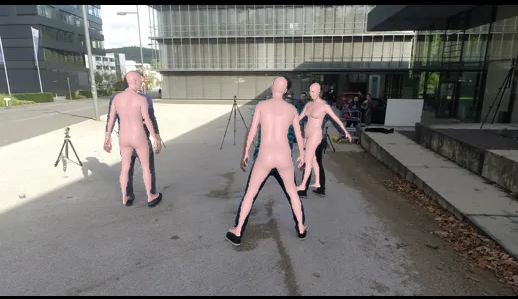} &
\mupotstrimmedgraphics{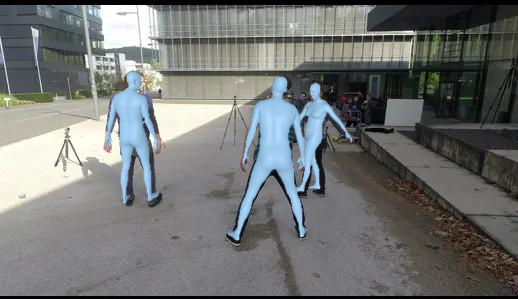} &
\mupotstrimmedgraphics{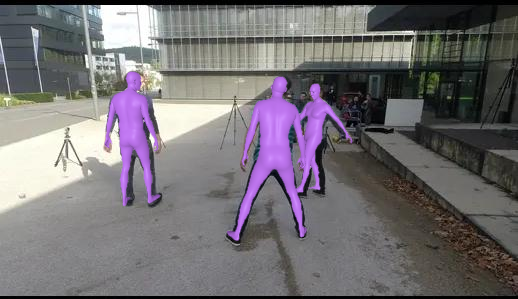} &
\mupotstrimmedgraphics{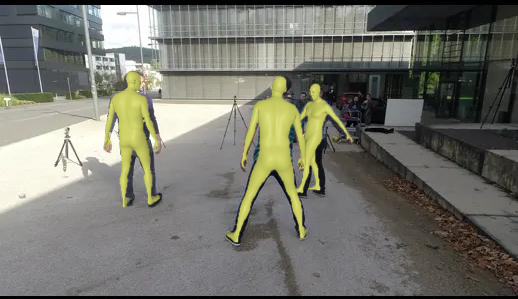} &
\includegraphics[height=2cm]{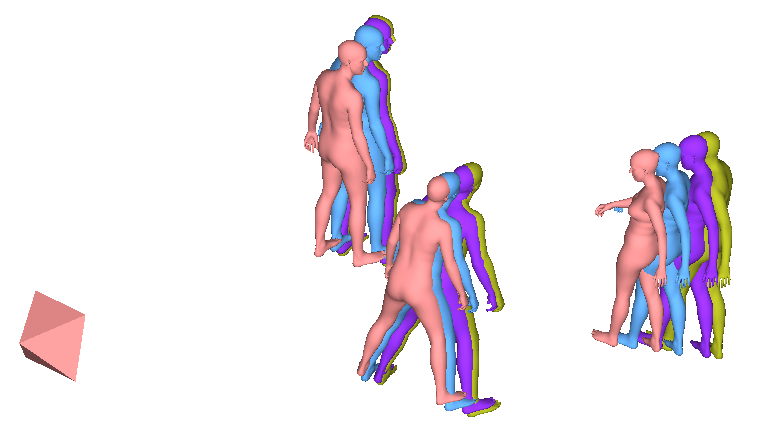} \\

\threedpwtrimmedgraphics{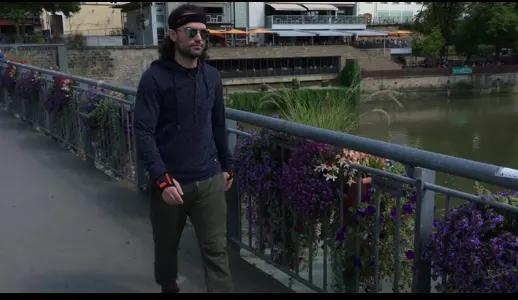} &
\threedpwtrimmedgraphics{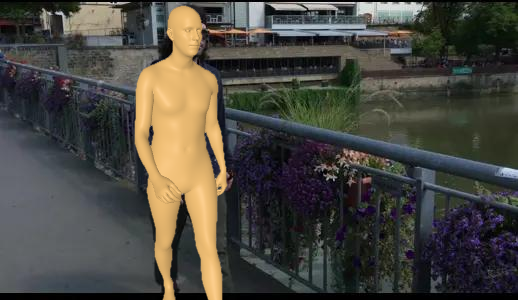} &
\threedpwtrimmedgraphics{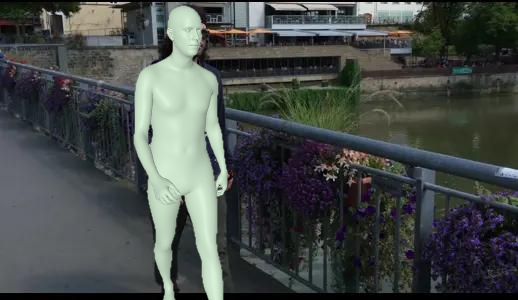} &
\threedpwtrimmedgraphics{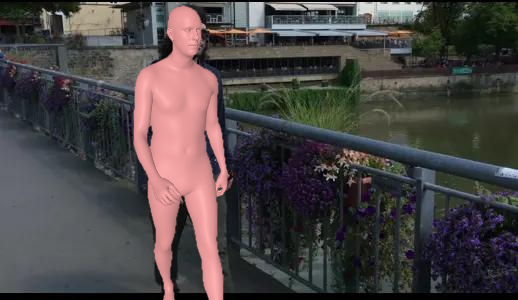} &
\threedpwtrimmedgraphics{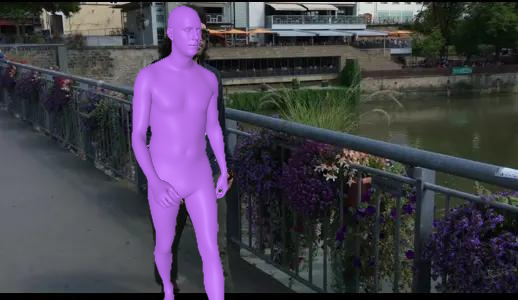} &
\includegraphics[height=2cm]{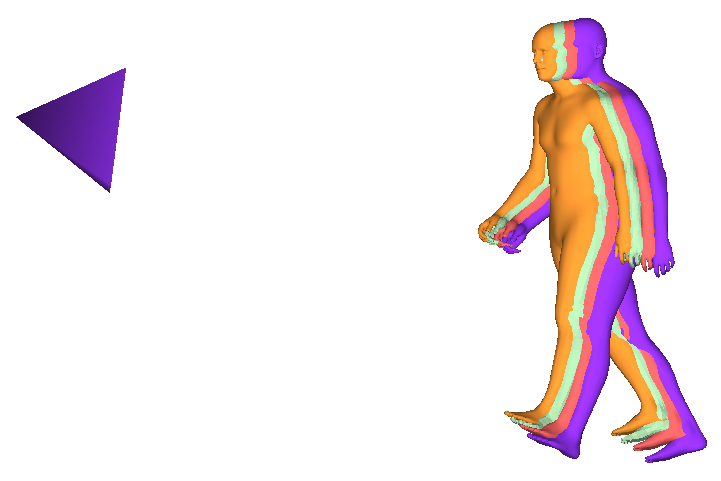} \\

\threedpwtrimmedgraphicsbis{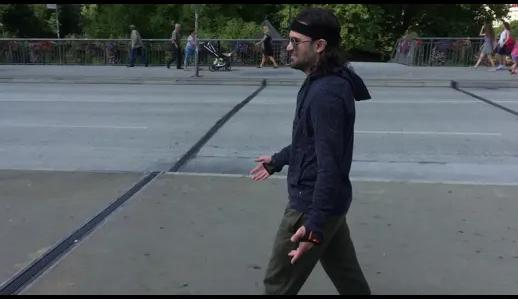} &
\threedpwtrimmedgraphicsbis{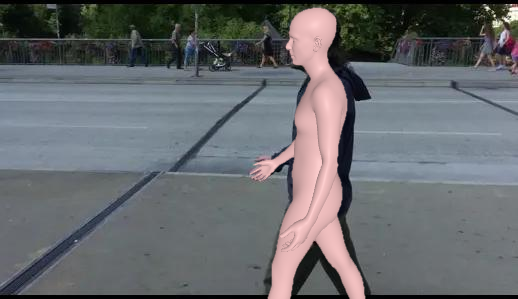} &
\threedpwtrimmedgraphicsbis{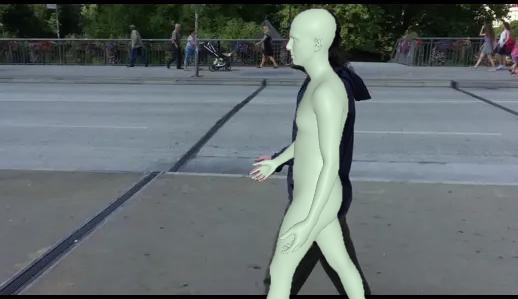} &
\threedpwtrimmedgraphicsbis{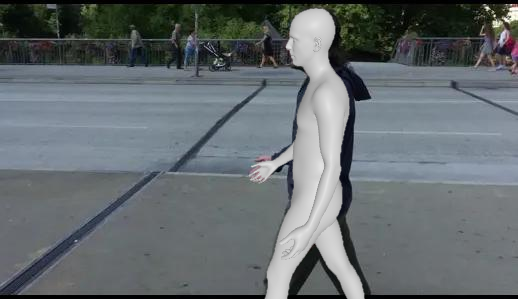} &
\threedpwtrimmedgraphicsbis{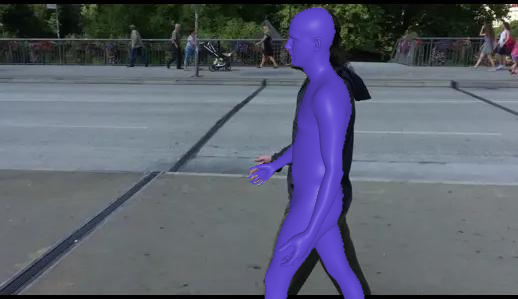} &
\includegraphics[height=2cm]{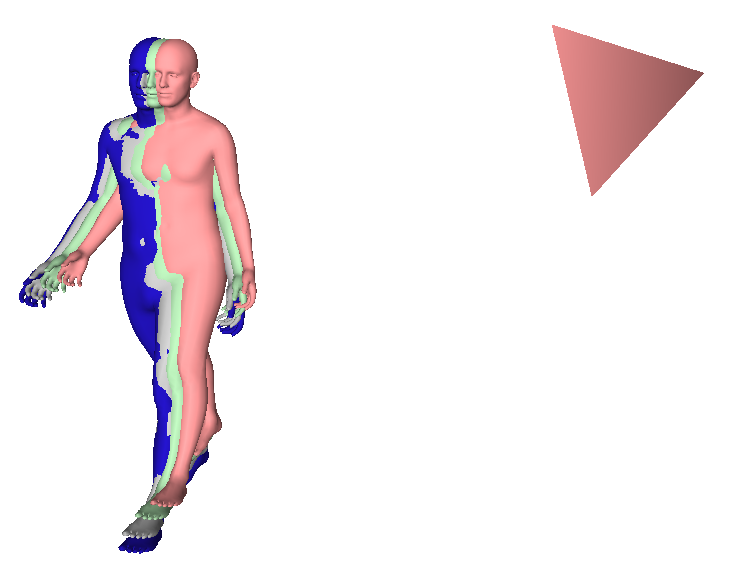} \\

\hifourdtrimmedgraphics{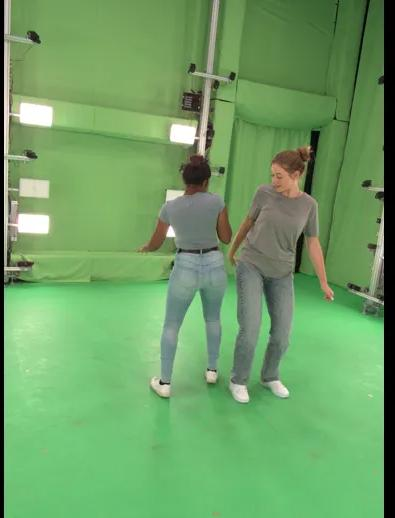} &
\hifourdtrimmedgraphics{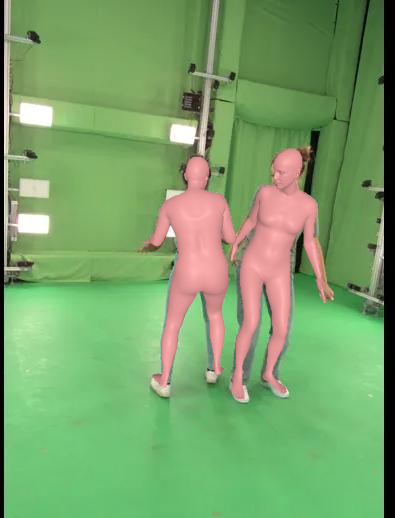} &
\hifourdtrimmedgraphics{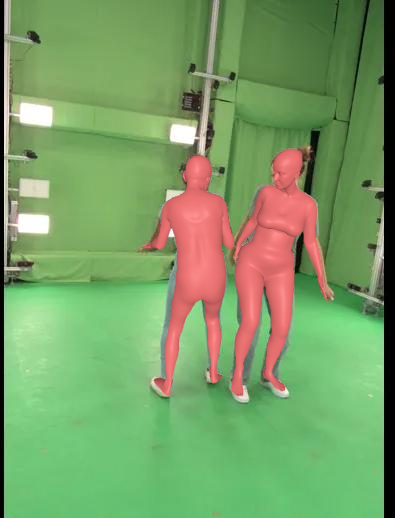} &
\hifourdtrimmedgraphics{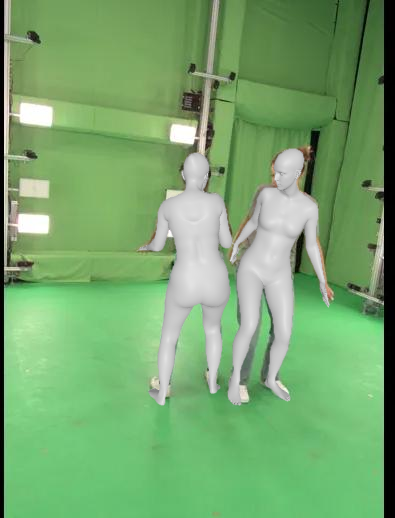} &
\hifourdtrimmedgraphics{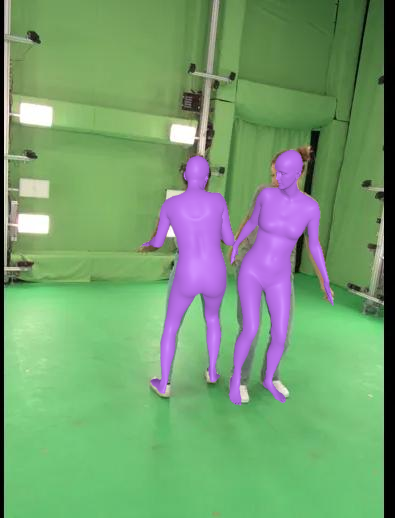} &
\includegraphics[height=2cm]{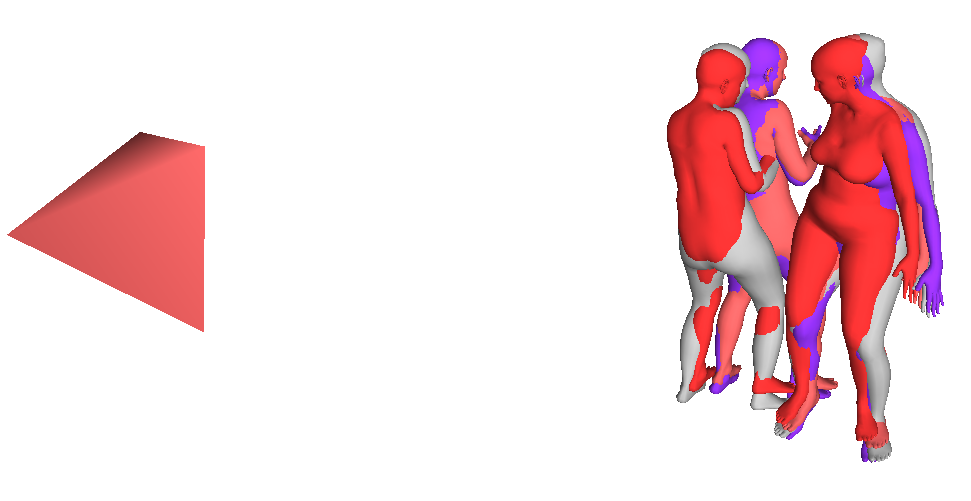} \\
\end{tabular}

\caption{Predictions assuming different body shapes.}
\label{fig:counterfactual_shape}
\end{subfigure}

\vspace{24pt}

\begin{subfigure}{\linewidth}
\centering
\begin{tabular}{cccccc}
\mupotstrimmedgraphics{figures/counterfactual/mupots_2807.jpg/scene0.png} &
\mupotstrimmedgraphics{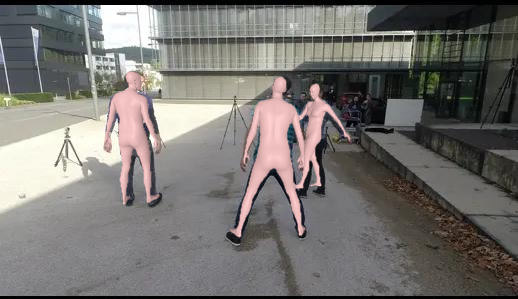} & 
\mupotstrimmedgraphics{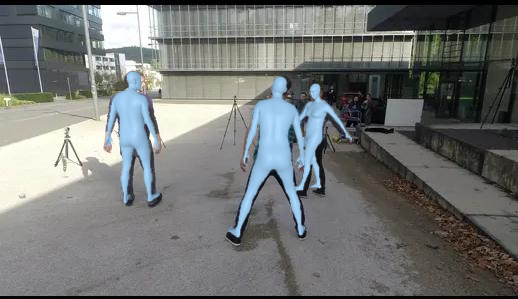} &
\mupotstrimmedgraphics{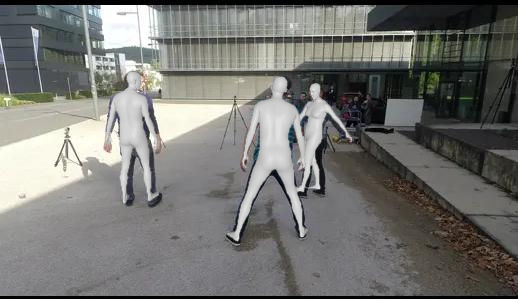} &
\mupotstrimmedgraphics{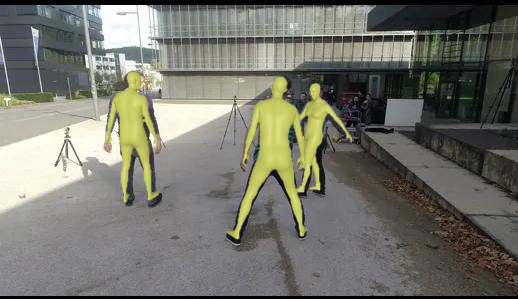} &
\includegraphics[height=2cm]{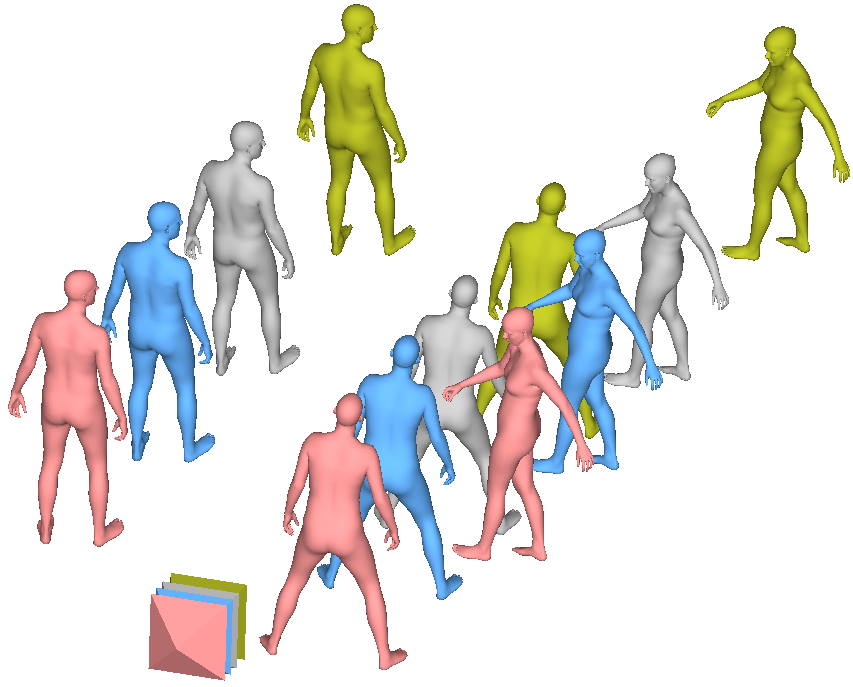} \\

\threedpwtrimmedgraphics{figures/counterfactual/3dpw_15346.jpg/scene0.png} &
\threedpwtrimmedgraphics{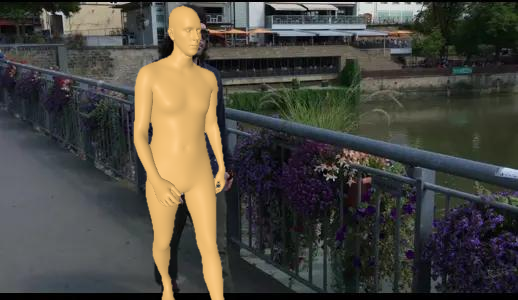} &
\threedpwtrimmedgraphics{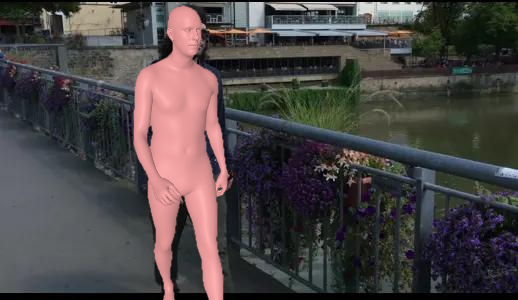} &
\threedpwtrimmedgraphics{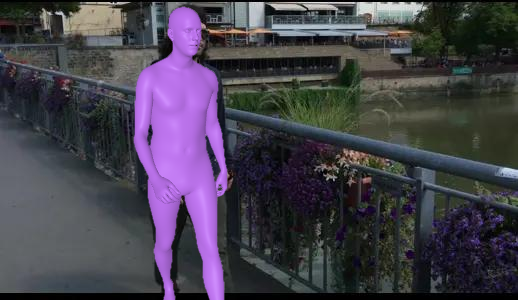} &
\threedpwtrimmedgraphics{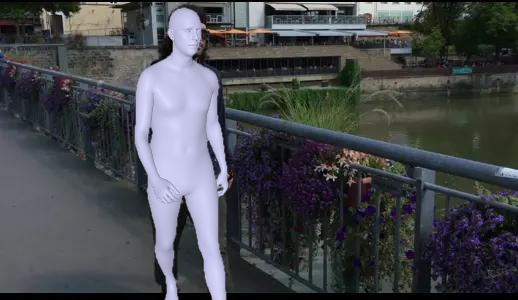} &
\includegraphics[height=2cm]{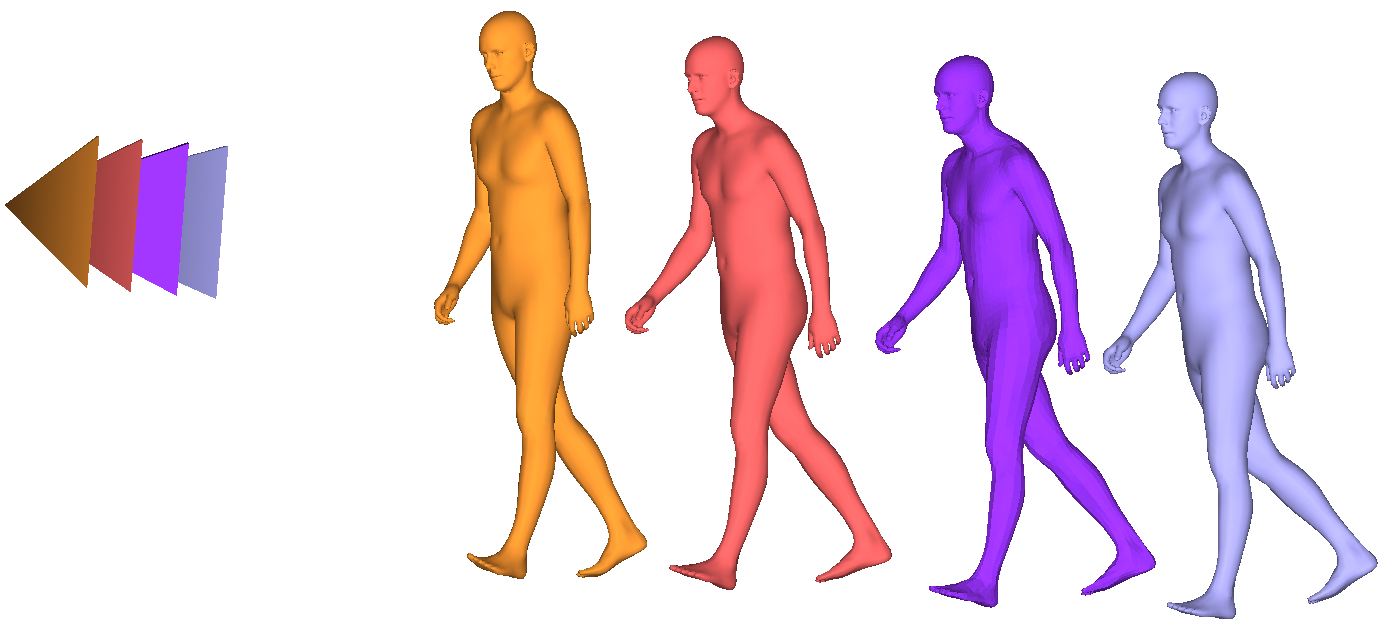} \\

\threedpwtrimmedgraphicsbis{figures/counterfactual/3dpw_14537.jpg/scene0.png} &
\threedpwtrimmedgraphicsbis{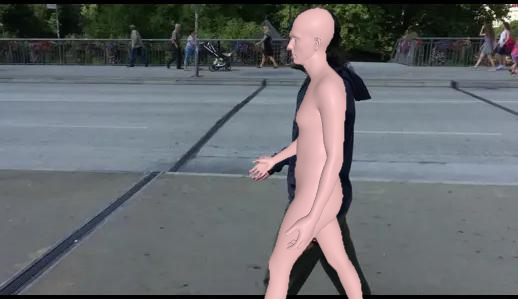} &
\threedpwtrimmedgraphicsbis{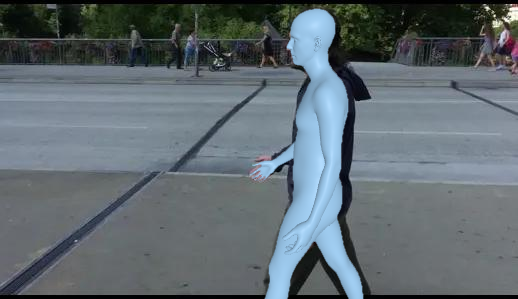} &
\threedpwtrimmedgraphicsbis{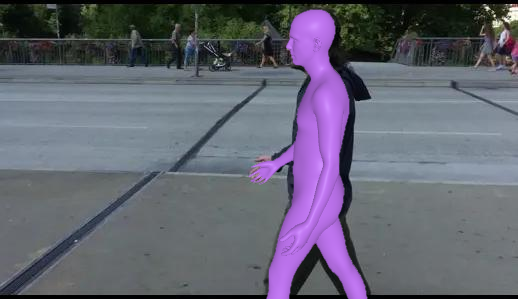} &
\threedpwtrimmedgraphicsbis{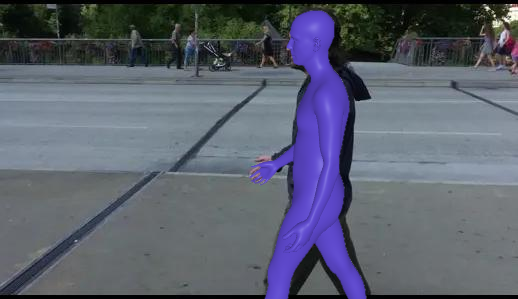} &
\includegraphics[height=2cm]{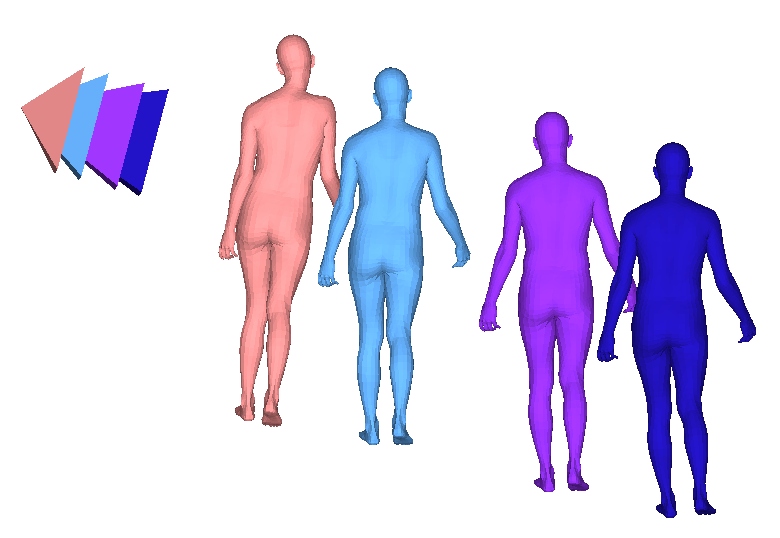} \\

\hifourdtrimmedgraphics{figures/counterfactual/hi4d_19639.jpg/scene0.png} &
\hifourdtrimmedgraphics{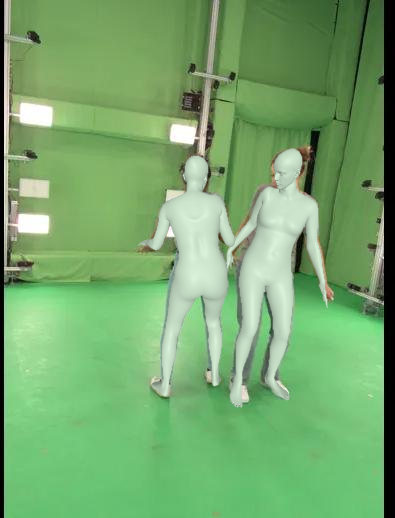} &
\hifourdtrimmedgraphics{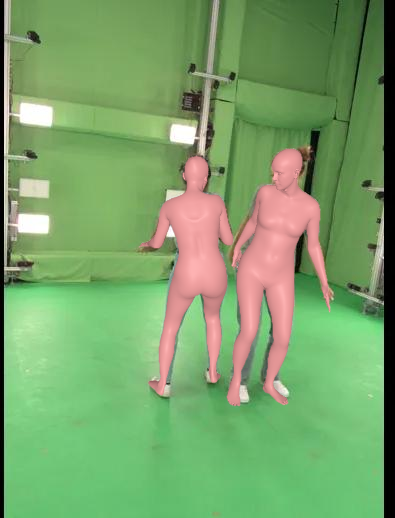} &
\hifourdtrimmedgraphics{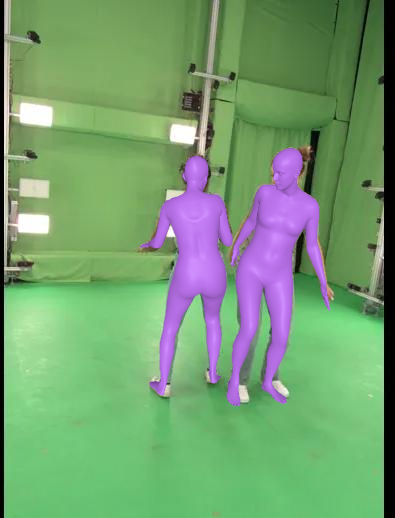} &
\hifourdtrimmedgraphics{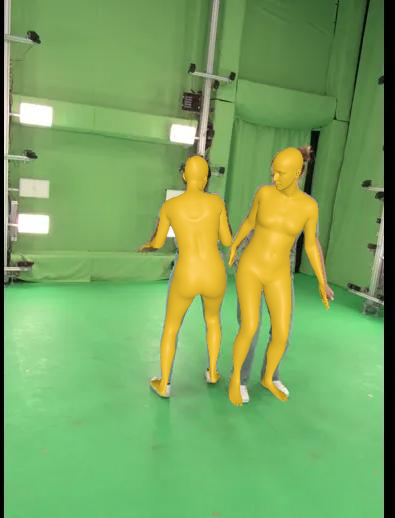} &
\includegraphics[height=2cm]{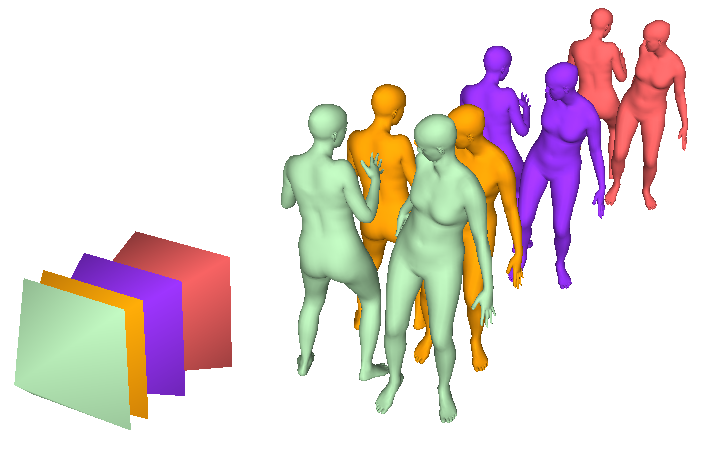} \\
\end{tabular}

\caption{Predictions assuming different focal lengths.}
\label{fig:counterfactual_focal}
\end{subfigure}

\caption{\textbf{Counterfactual experiments using different external inputs.} Input image (left) and predictions using different external inputs visualized from camera (middle) and side view (right).}
\label{fig:counterfactual_examples}
\end{figure*}

In addition to the numerical results reported in the main paper, Fig.~\ref{fig:counterfactual_examples} provides qualitative results of counterfactual experiments that illustrate the ability of our approach to model dependencies between attributes in the mesh recovery problem. In Fig.~\ref{fig:counterfactual_shape}, we vary the principal component of body shape parameters (as external inputs) while keeping  camera intrinsics constant, and observe the effect on predicted distances to the camera.
Similarly,  Fig.~\ref{fig:counterfactual_focal} illustrates the effect of setting different focal lengths as inputs,
demonstrating how this variation influences other variables, particularly the distance to the camera. %

\section{Uncertainty modeling}
\label{sec:uncertainty}
\begin{figure*}
    \centering
    \begin{subfigure}{\linewidth}
        \includegraphics[width=\linewidth]{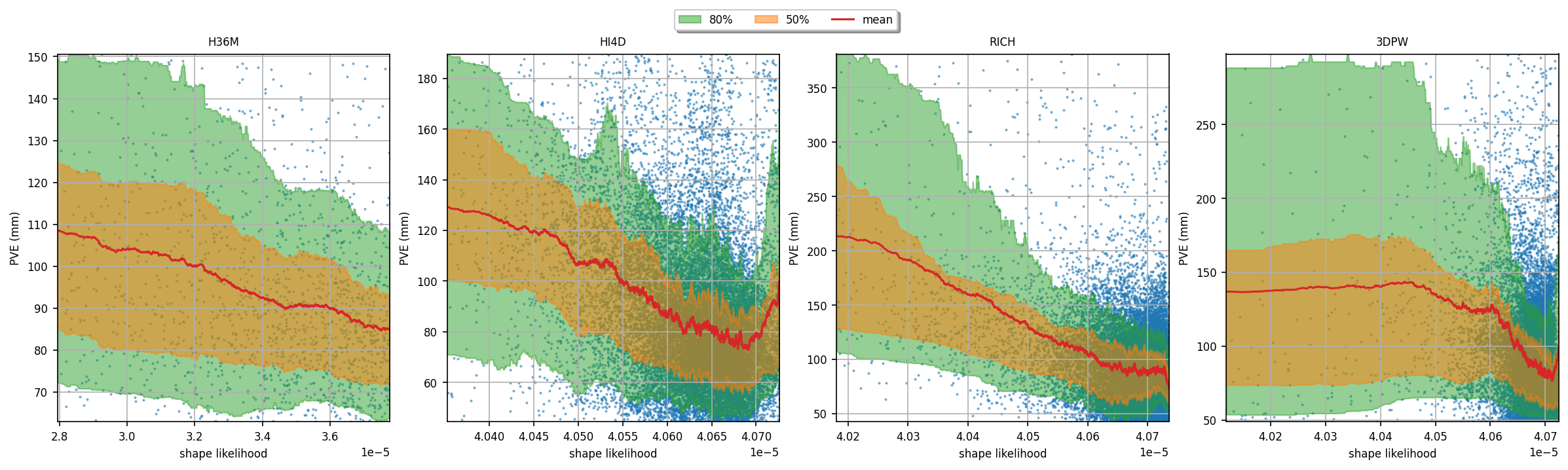}    
        \caption{Shape likelihood.%
        }
    \end{subfigure}
    \begin{subfigure}{\linewidth}
        \includegraphics[width=\linewidth]{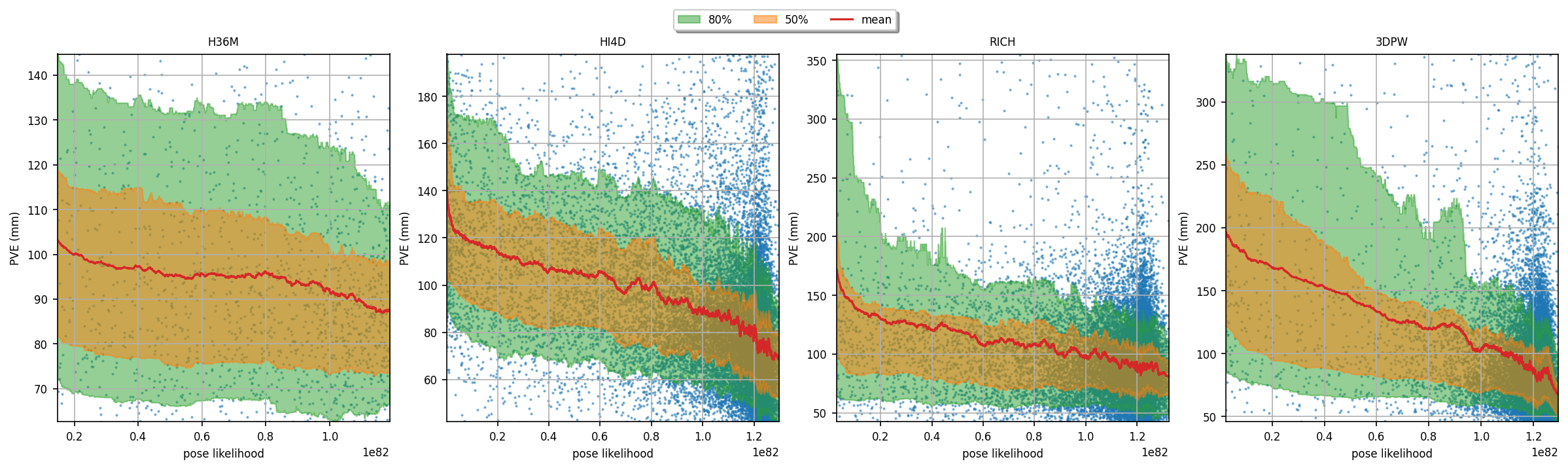}
        \caption{Pose likelihood.%
        }
    \end{subfigure}
    \begin{subfigure}{\linewidth}
        \includegraphics[width=\linewidth]{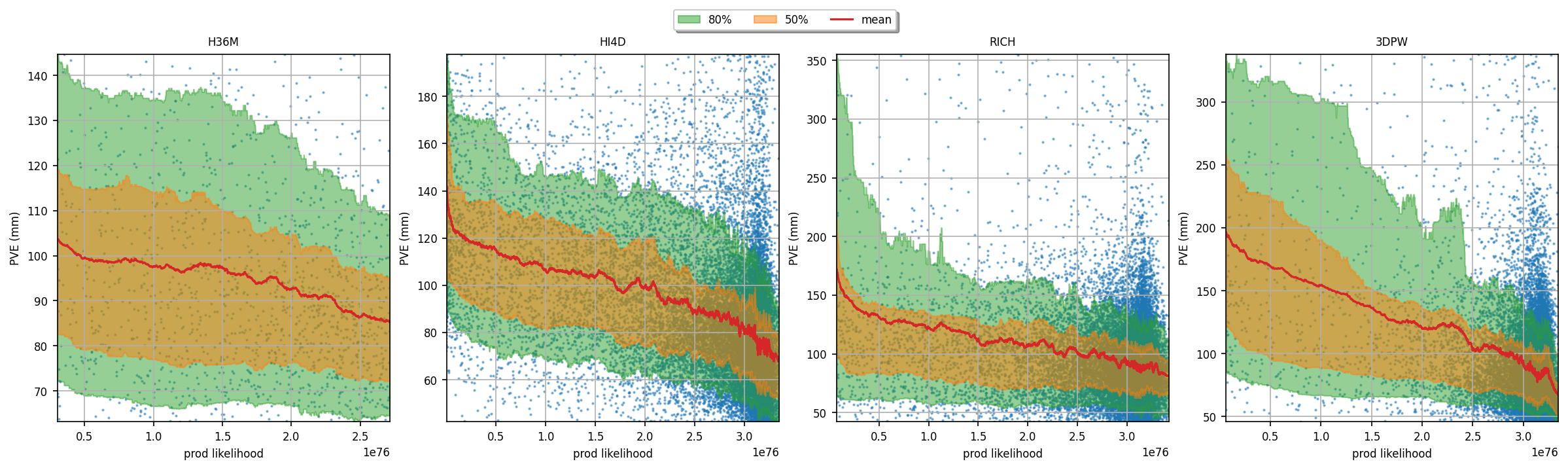}
        \caption{Joint likelihood $p(\hat{t}, \hat{\theta}, \hat{\beta} \vert \mathcal{I})$.}
        \label{subfig:prod_likelihood_vs_error}
    \end{subfigure}    
    \caption{
    \textbf{Relationship between prediction error and predicted likelihood across datasets.}
    The predicted likelihood values are correlated with the test error, providing a proxy for prediction confidence.
    Trend curves (shown in red, yellow, and green) are calculated using a sliding window of 400 samples.
    }
    \label{fig:likelihood_vs_error}
\end{figure*}

Empirically, we observe a correlation between the conditional likelihoods of our predictions -- \ie{} the value of conditional probability densities predicted by our Bayesian network -- and actual prediction errors, as illustrated in Fig.~\ref{fig:likelihood_vs_error} for various test sets.
This suggests that the proposed model is able to capture the uncertainty of its predictions to some extent, which could be useful in downstream applications.

\section{Failure cases}
\label{sec:failure_cases}
\begin{figure*}
    \small
    \centering
    \setlength{\tabcolsep}{3pt}
	\newcommand{\mywildcolumn}[1]{\includegraphics[width=4.1cm,height=3cm,keepaspectratio]{figures/wild_results/#1/input0.jpg} & \includegraphics[width=4.1cm,height=3cm,keepaspectratio]{figures/wild_results/#1/detection0.jpg} & \includegraphics[width=4.1cm,height=3cm,keepaspectratio]{figures/wild_results/#1/prediction0_.jpg} & \includegraphics[width=4.1cm,height=2.5cm,keepaspectratio]{figures/wild_results/#1/side_view.png}}
    \begin{tabular}{cccc}
        \textbf{Input} & \textbf{Ref. keypoint detection} & \textbf{Prediction} & \textbf{Side-view} \\
        \mywildcolumn{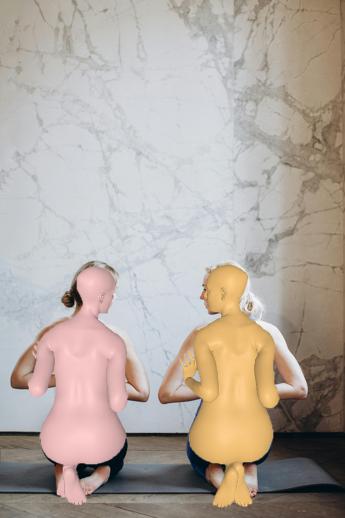} \\
       \mywildcolumn{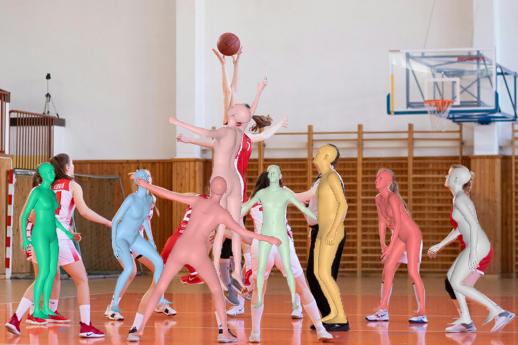} \\
    \end{tabular}
    \caption{\label{fig:limitations}\textbf{Limitations}. 
    Like other existing approaches, our method struggles with unusual poses far from the training data (top row).
    Images depicting multiple person with reference keypoints (head) reprojecting at similar 2D locations can lead to missed detections and ambiguous predictions (bottom row).}
\end{figure*}

Overall, \Ours{} produces plausible predictions.
However, it also inherits common limitations of existing mesh recovery methods. 
Notable failure cases (not specific to our method) include unusual poses that deviate significantly from the training data (Fig.~\ref{fig:limitations} top row). 
Additionally, scenes with mutually occluding persons introduce ambiguity in the detection task (Fig.~\ref{fig:limitations} bottom row).

\section{Bayesian network connectivity}
\label{sec:connectivity_ablation}
\begin{figure}
    \centering
    \footnotesize
    \begin{subfigure}{0.30\textwidth}
    \centering
    \includegraphics[trim={1.1cm 1.1cm 1.1cm 1cm},clip,scale=0.24]{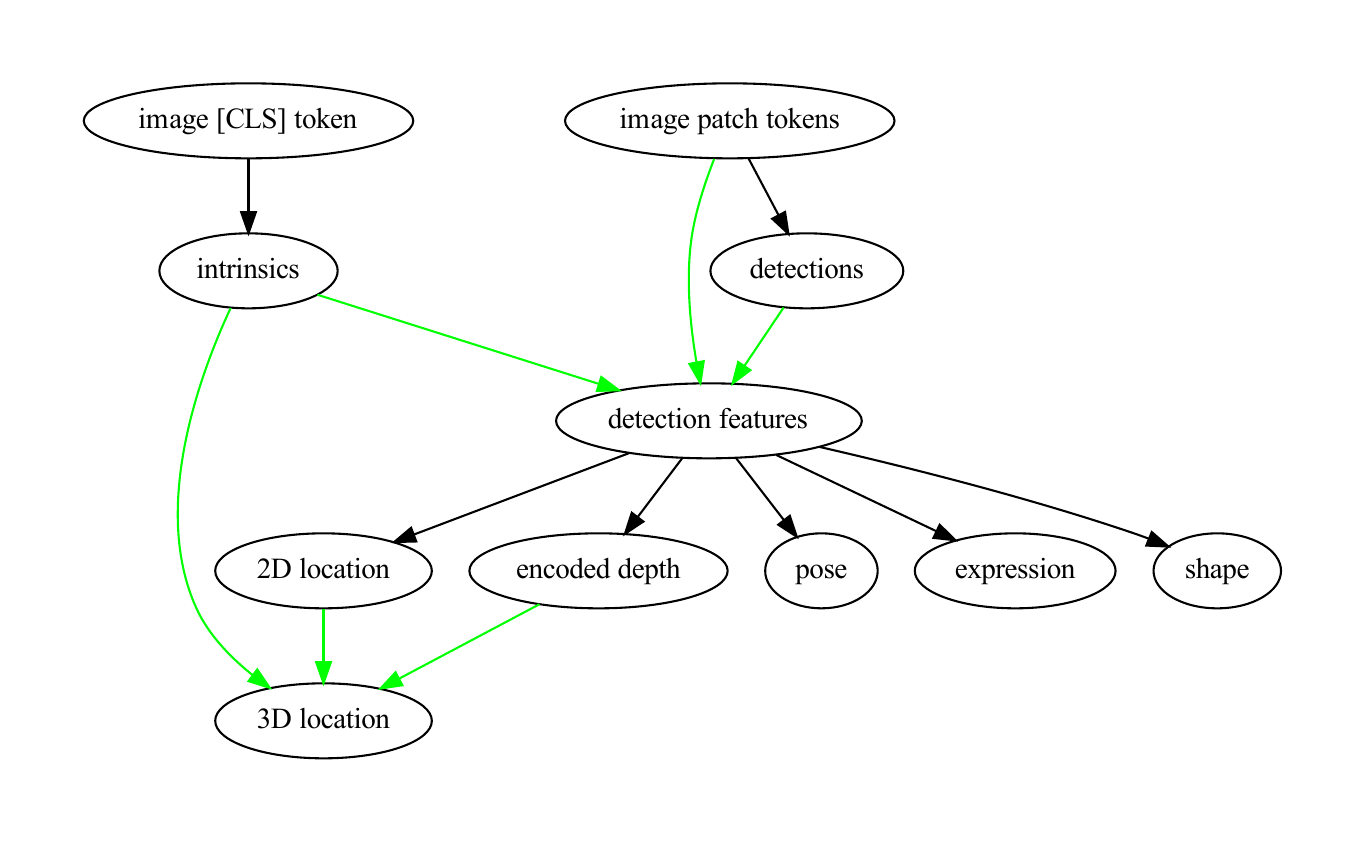}
    \caption{\emph{Naive Bayes}.}
    \end{subfigure}
    \begin{subfigure}{0.25\textwidth}
    \centering
    \includegraphics[trim={1.1cm 1.1cm 1.1cm 1.1cm},clip,scale=0.24]{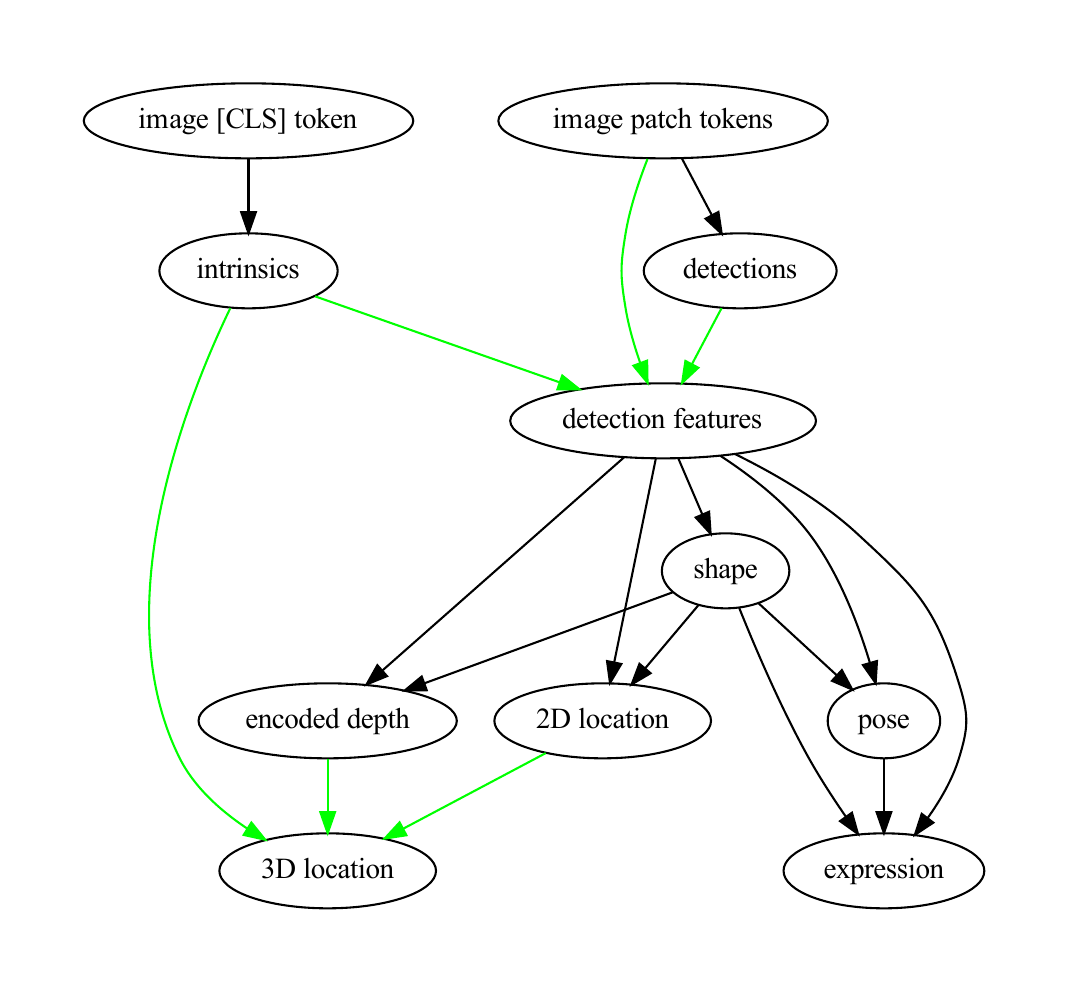}
    \caption{\emph{\Ours} (default).}
    \end{subfigure}
    \begin{subfigure}{0.24\textwidth}
    \centering
    \includegraphics[trim={1.1cm 1.1cm 1.1cm 1.1cm},clip,scale=0.24]{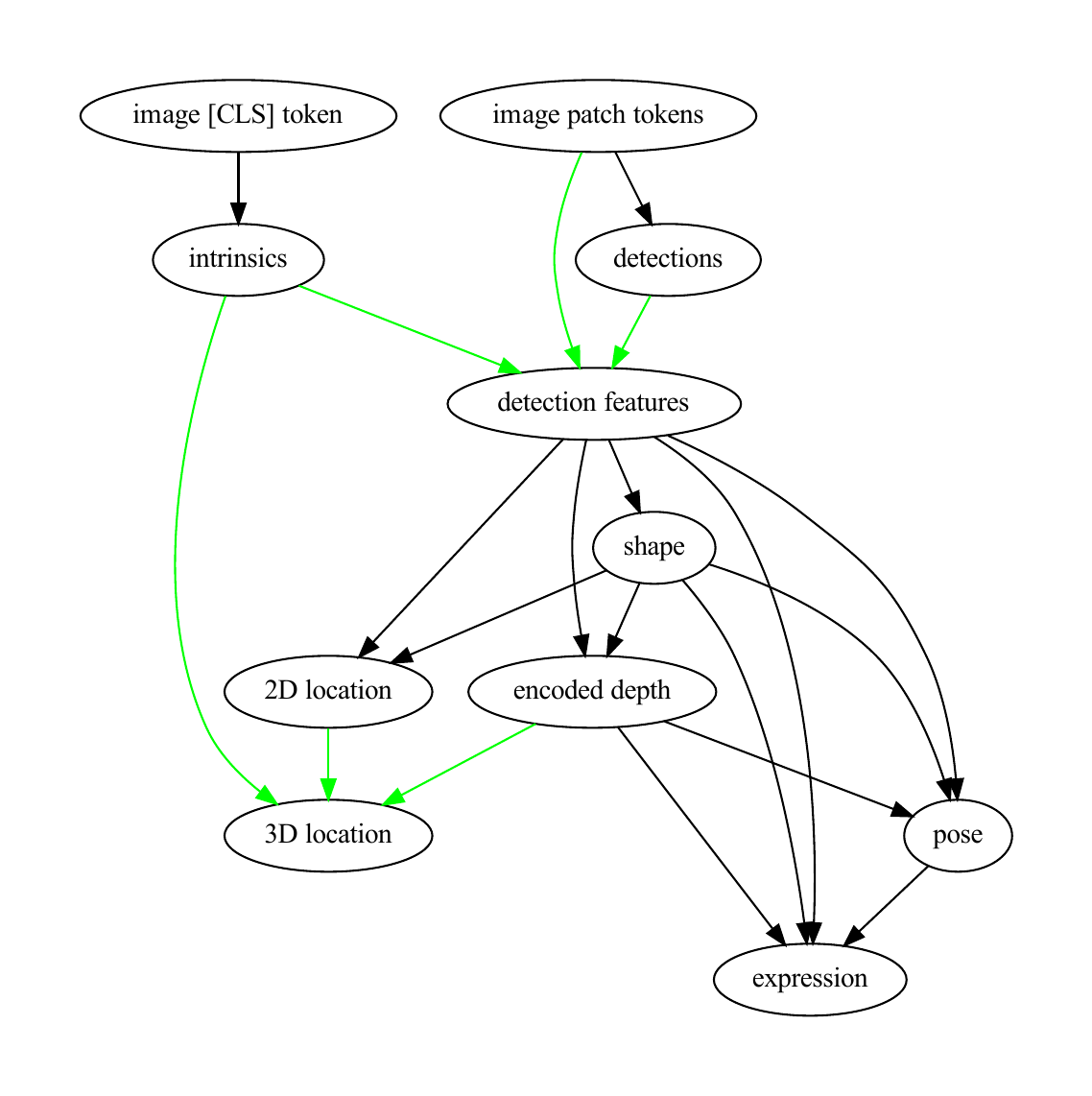}
    \caption{\emph{Variant1}.}
    \end{subfigure}
    \begin{subfigure}{0.21\textwidth}
    \centering
    \includegraphics[trim={1.1cm 1.1cm 1.1cm 1.1cm},clip,scale=0.24]{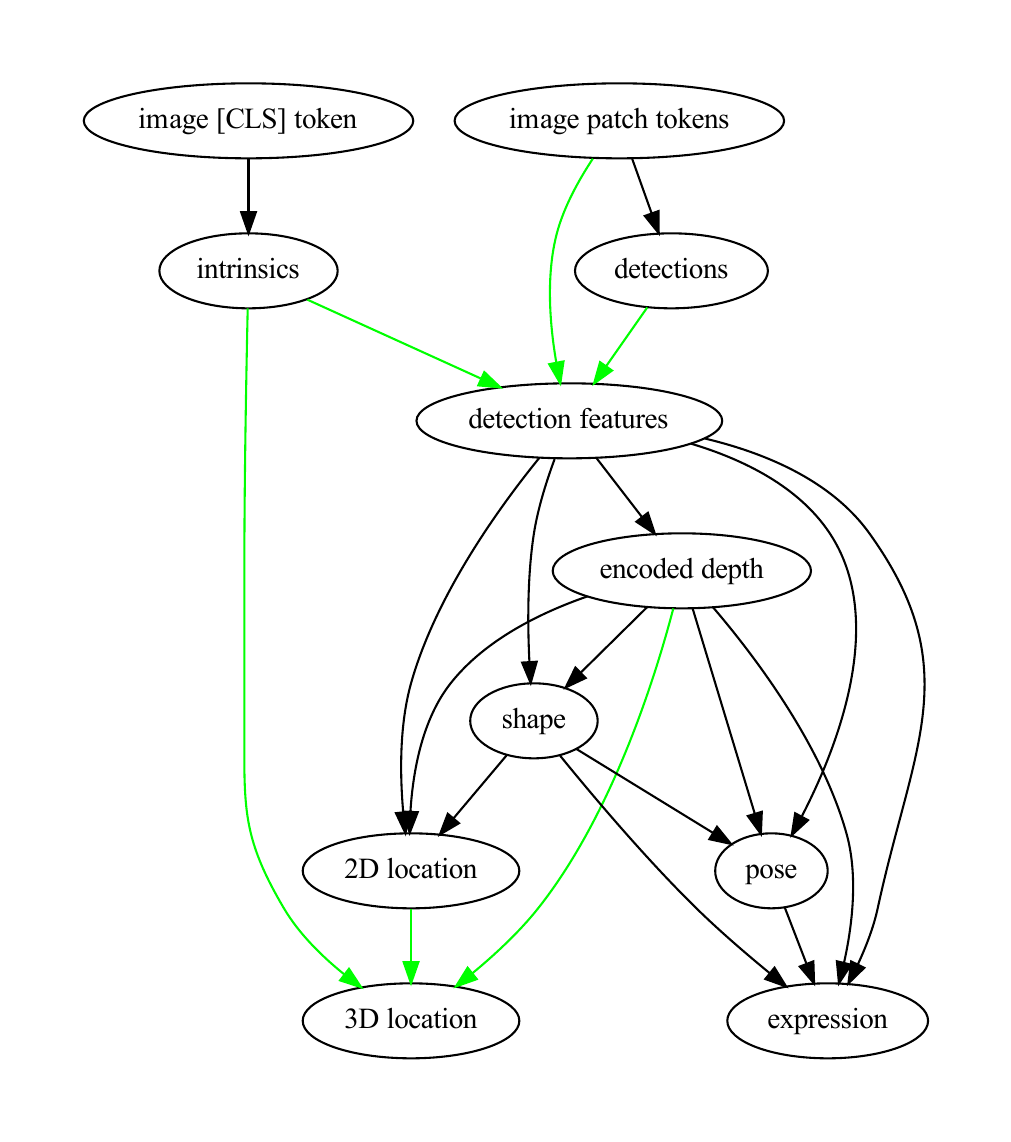}
    \caption{\emph{Variant2}.}
    \end{subfigure}
    \caption{\textbf{Connectivity of different Bayesian networks considered for this study.} Deterministic dependencies between variables are represented in green.}
    \label{fig:graph_connectivity}
\end{figure}

Beyond \emph{\Ours} and the \emph{Naive Bayes} baseline presented in the main paper, we also experimented with two additional variants to study the impact of Bayesian network connectivity on numerical performances. Fig.~\ref{fig:graph_connectivity} shows the full connectivity of the different Bayesian networks considered in this study, from which the graphical model Fig.~1 of the main paper is extracted. 
\emph{Variant1} features a denser set of conditional dependency connections compared to \emph{\Ours}, and in \emph{Variant2} the dependency order between body shape and encoded depth variables is furthermore permuted. We report results of quantitative evaluations in Table~\ref{tab:eval_different_connectivity}.
The dependency order in \emph{Variant2} prevents from properly exploiting external camera intrinsics and body shape inputs, leading to much larger absolute position errors in this setting than with \emph{\Ours} and \emph{Variant1}, but still outperforming the \emph{Naive Bayes} baseline (\eg{} on \emph{Human3.6M} in  \emph{Single-View intr-shape} setup, $PE=676.9mm$ for \emph{Variant2} \vs $284.6mm$ for \emph{\Ours}, $366.2mm$ for \emph{Variant1}, and $898.0mm$ for \emph{Naive Bayes}).
Overall, \emph{\Ours} achieves better numerical performances than \emph{Variant1}. The restricted connectivity of \emph{\Ours} imposes stronger inductive biases than the connectivity of \emph{Variant1}, and we posit it helps learning meaningful correlations between human attributes. This observation is arguably dependent of our training strategy, and we expect that further increasing the amount and variability of training data would diminish the benefits of imposing such inductive priors.%

\begin{table*}
\caption{\textbf{Quantitative evaluation} for different Bayesian network connectivity settings, with different additional inputs: camera intrinsics (\emph{intr}), distance to the camera (\emph{dist}), or known body shape (\emph{shape}).}
\label{tab:eval_different_connectivity}
\centering
\footnotesize
\setlength{\tabcolsep}{2pt}
    \centering
    \scriptsize
\setlength{\tabcolsep}{2pt}
\begin{tabular}{lcl|ccc|ccc|ccc|ccc|cc}
\toprule
\multicolumn{3}{l|}{\bf{Experiments}} & \multicolumn{3}{c|}{\bf{Human3.6M}}  & \multicolumn{3}{c|}{\bf{HI4D}} & \multicolumn{3}{c|}{\bf{RICH}} & \multicolumn{3}{c|}{\textbf{3DPW}} & \multicolumn{2}{c}{\textbf{MuPOTS}} \\
& Ext. input & Connectivity & PVE$\downarrow$  & PA-PVE$\downarrow$  & PE$\downarrow$  &  PVE$\downarrow$  & PA-PVE$\downarrow$  & PE$\downarrow$  & PVE$\downarrow$  & PA- PVE$\downarrow$  & PE$\downarrow$ & PJE$\downarrow$ & PA-PJE$\downarrow$ & PE$\downarrow$ & PCK-Matched$\uparrow$ & PCK-All$\uparrow$\\
\midrule
\parbox[t]{2mm}{\multirow{20}{*}{\rotatebox[origin=c]{90}{Single-View}}} & \multirow{4}{*}{none} & Naive-Bayes       & 113.6 & 55.3 & \textbf{808.5} & 92.8 & \textbf{47.7} & 168.7 & 119.6 & \textbf{46.4} & 775.3 & 73.9 & 47.9 & \textbf{662.6} & 84.8 & 73.4 \\
&  & \Ours            & \textbf{98.9} & \textbf{54.3} & 1145.2 & \textbf{81.4} & 48.0 & \textbf{144.7} & \textbf{106.5} & 46.6 & 785.1 & \textbf{69.2} & \textbf{46.4} & 728.9 & \textbf{85.2} & 74.5 \\
&  & Variant1          & 103.9 & 58.4 & 1221.0 & 91.2 & 49.0 & 268.8 & 130.0 & 49.8 & 710.3 & 71.2 & 47.4 & 837.5 & 85.1 & 74.5 \\
&  & Variant2          & 100.6 & 55.1 & 1227.8 & 101.5 & \textbf{47.7} & 182.1 & 127.5 & 46.6 & \textbf{612.2} & 70.2 & 47.0 & 908.2 & 83.9 & \textbf{74.6} \\
\cmidrule{2-17}
& \multirow{4}{*}{intr} & Naive-Bayes & 104.0 & 54.1 & 904.3 & 93.3 & 47.7 & 391.3 & 118.0 & \textbf{46.5} & 1060.5 & 73.8 & 47.7 & 428.4 & 84.0 & 72.7 \\
&  & \Ours      & \textbf{94.3} & \textbf{53.9} & 648.2 & \textbf{83.0} & 48.0 & 305.3 & \textbf{106.6} & 46.7 & \textbf{972.3} & \textbf{69.5} & \textbf{46.4} & 337.2 & 84.7 & 74.0 \\
&  & Variant1    & 100.3 & 57.5 & \textbf{646.8} & 90.1 & 48.6 & \textbf{286.7} & 129.3 & 49.8 & 1029.9 & 71.0 & 47.2 & \textbf{298.0} & \textbf{85.4} & \textbf{74.8} \\
&  & Variant2    & 95.4 & 54.2 & 682.8 & 101.0 & \textbf{47.5} & 394.5 & 126.1 & 46.5 & 1049.6 & 70.0 & 46.9 & 365.6 & 83.7 & 74.5 \\
\cmidrule{2-17}
& \multirow{4}{*}{intr-dist} & Naive-Bayes & 104.0 & 54.0 & 98.1 & 93.3 & 47.7 & 83.3 & 118.0 & \textbf{46.4} & 115.1 & 81.2 & 52.6 & 112.8 & -- & -- \\
&  & \Ours & \textbf{94.3} & \textbf{53.9} & 89.4 & \textbf{83.0} & 48.0 & \textbf{69.7} & \textbf{106.6} & 46.7 & \textbf{99.1} & 76.4 & 51.2 & 104.1 & -- & -- \\
&  & Variant1 & 100.8 & 57.4 & 90.1 & 90.1 & 48.6 & 77.6 & 129.7 & 50.0 & 124.4 & 77.5 & 51.5 & \textbf{101.3} & -- & -- \\
&  & Variant2 & 95.4 & 54.2 & \textbf{88.5} & 100.8 & \textbf{47.4} & 90.7 & 126.1 & 46.6 & 114.8 & \textbf{75.8} & \textbf{51.0} & 106.3 & -- & -- \\
\cmidrule{2-15}
& \multirow{4}{*}{intr-shape} & Naive-Bayes & 73.6 & 54.1 & 898.0 & 70.9 & 47.3 & 385.0 & 84.2 & 47.9 & 1055.2 & 73.5 & 50.0 & 437.8 & -- & -- \\
&  & \Ours & \textbf{70.3} & \textbf{53.7} & \textbf{284.6} & \textbf{62.8} & 47.6 & 132.2 & \textbf{82.3} & \textbf{48.1} & \textbf{417.3} & \textbf{69.9} & \textbf{48.7} & 354.2 & -- & -- \\
&  & Variant1 & 80.2 & 57.4 & 366.2 & 62.9 & 48.2 & \textbf{113.9} & 87.5 & 51.3 & 508.7 & 72.4 & 49.6 & \textbf{267.9} & -- & -- \\
&  & Variant2 & 72.8 & 54.5 & 676.9 & 65.4 & \textbf{47.1} & 382.0 & \textbf{82.3} & 48.1 & 1038.8 & 70.4 & 49.3 & 375.5 & -- & -- \\
\cmidrule{2-15}
& \multirow{4}{*}{intr-shape-dist} & Naive-Bayes & 73.6 & 54.1 & 56.4 & 70.8 & 47.3 & 59.0 & 84.2 & \textbf{47.8} & 76.8 & 75.9 & 51.0 & 101.6 & -- & -- \\
&  & \Ours & \textbf{70.3} & \textbf{53.7} & \textbf{56.1} & \textbf{62.8} & 47.6 & \textbf{46.2} & \textbf{82.3} & 48.1 & 73.5 & \textbf{72.0} & \textbf{49.6} & 99.8 & -- & -- \\
&  & Variant1 & 80.5 & 57.4 & 65.2 & \textbf{62.8} & 48.2 & 46.4 & 88.0 & 51.4 & 77.9 & 73.9 & 50.4 & \textbf{97.5} & -- & -- \\
&  & Variant2 & 73.1 & 54.5 & 57.9 & 65.5 & \textbf{47.1} & 51.6 & 82.9 & 48.2 & \textbf{70.2} & \textbf{72.0} & 50.2 & \textbf{97.5} & -- & -- \\
\cmidrule{1-15}
\multirow{20}{*}{\rotatebox[origin=c]{90}{Multi-View}} & \multirow{4}{*}{none} & Naive-Bayes       & 104.8 & 43.8 & \textbf{828.8} & 85.0 & \textbf{35.3} & 162.2 & 105.8 & 37.1 & 770.5 & -- & -- & -- & -- & -- \\
& & \Ours            & 90.6 & \textbf{42.6} & 1148.3 & \textbf{75.2} & 35.5 & \textbf{142.9} & \textbf{93.0} & \textbf{36.2} & 724.5 & -- & -- & -- & -- & -- \\
& & Variant1          & 96.0 & 45.3 & 1214.8 & 83.4 & \textbf{35.3} & 275.6 & 113.4 & 36.8 & 622.8 & -- & -- & -- & -- & -- \\
& & Variant2          & \textbf{88.4} & 42.7 & 1240.7 & 94.7 & \textbf{35.3} & 181.8 & 115.4 & 36.5 & \textbf{613.0} & -- & -- & -- & -- & -- \\
\cmidrule{2-12}
& \multirow{4}{*}{intr} & Naive-Bayes & 98.9 & 43.6 & 895.9 & 85.6 & 35.4 & 381.8 & 104.3 & 37.1 & 1057.1 & -- & -- & -- & -- & -- \\
& & \Ours      & \textbf{88.8} & 42.6 & 679.1 & \textbf{77.0} & 35.5 & 308.9 & \textbf{92.8} & \textbf{36.2} & \textbf{903.4} & -- & -- & -- & -- & -- \\
& & Variant1    & 94.8 & 45.2 & 668.9 & 82.8 & \textbf{35.3} & \textbf{281.0} & 112.8 & 36.8 & 930.7 & -- & -- & -- & -- & -- \\
& & Variant2    & 84.8 & \textbf{42.4} & 678.8 & 94.4 & \textbf{35.3} & 385.4 & 114.1 & 36.5 & 1050.0 & -- & -- & -- & -- & -- \\
\cmidrule{2-12}
& \multirow{4}{*}{intr-dist} & Naive-Bayes & 98.9 & 43.6 & 96.8 & 85.5 & 35.4 & 78.7 & 104.3 & 37.1 & 103.4 & -- & -- & -- & -- & -- \\
& & \Ours & 88.8 & 42.6 & 90.3 & 77.0 & 35.5 & \textbf{68.8} & \textbf{92.8} & \textbf{36.2} & \textbf{89.8} & -- & -- & -- & -- & -- \\
& & Variant1 & 95.4 & 45.4 & 90.4 & \textbf{82.9} & \textbf{35.3} & 75.7 & 113.5 & 37.0 & 115.2 & -- & -- & -- & -- & -- \\
& & Variant2 & \textbf{84.9} & \textbf{42.5} & \textbf{81.6} & 94.3 & \textbf{35.3} & 88.2 & 114.2 & 36.7 & 107.8 & -- & -- & -- & -- & -- \\
\cmidrule{2-12}
& \multirow{4}{*}{intr-shape} & Naive-Bayes & 67.3 & 43.9 & 890.2 & 63.1 & \textbf{34.9} & 376.3 & 80.3 & 39.2 & 1053.8 & -- & -- & -- & -- & -- \\
& & \Ours & \textbf{62.8} & 42.9 & \textbf{275.0} & 57.1 & 35.2 & 136.9 & \textbf{77.2} & \textbf{38.6} & \textbf{439.0} & -- & -- & -- & -- & -- \\
& & Variant1 & 71.8 & 45.4 & 350.7 & \textbf{55.8} & 35.0 & \textbf{116.0} & 82.3 & 39.3 & 518.2 & -- & -- & -- & -- & -- \\
& & Variant2 & 65.2 & \textbf{42.8} & 673.7 & 58.3 & 35.1 & 373.6 & 78.9 & 38.9 & 1040.2 & -- & -- & -- & -- & -- \\
\cmidrule{2-12}
& \multirow{4}{*}{intr-shape-dist} & Naive-Bayes & 67.3 & 43.9 & 55.8 & 63.0 & \textbf{34.9} & 54.3 & 80.3 & 39.2 & 75.5 & -- & -- & -- & -- & -- \\
& & \Ours & \textbf{62.8} & \textbf{42.9} & \textbf{54.2} & 57.1 & 35.2 & 45.7 & \textbf{77.2} & \textbf{38.6} & 74.1 & -- & -- & -- & -- & -- \\
& & Variant1 & 72.2 & 45.5 & 62.5 & \textbf{55.8} & 35.1 & \textbf{45.0} & 83.0 & 39.5 & 80.5 & -- & -- & -- & -- & -- \\
& & Variant2 & 65.5 & \textbf{42.9} & 56.5 & 58.7 & 35.1 & 48.4 & 79.7 & 39.1 & \textbf{72.2} & -- & -- & -- & -- & -- \\
\bottomrule
\end{tabular}
\end{table*}

\section{Implementation details}
\label{sec:implementation_details}
\subsection{Additional synthetic data}
\begin{figure}
    \centering
    \includegraphics[width=0.4\linewidth]{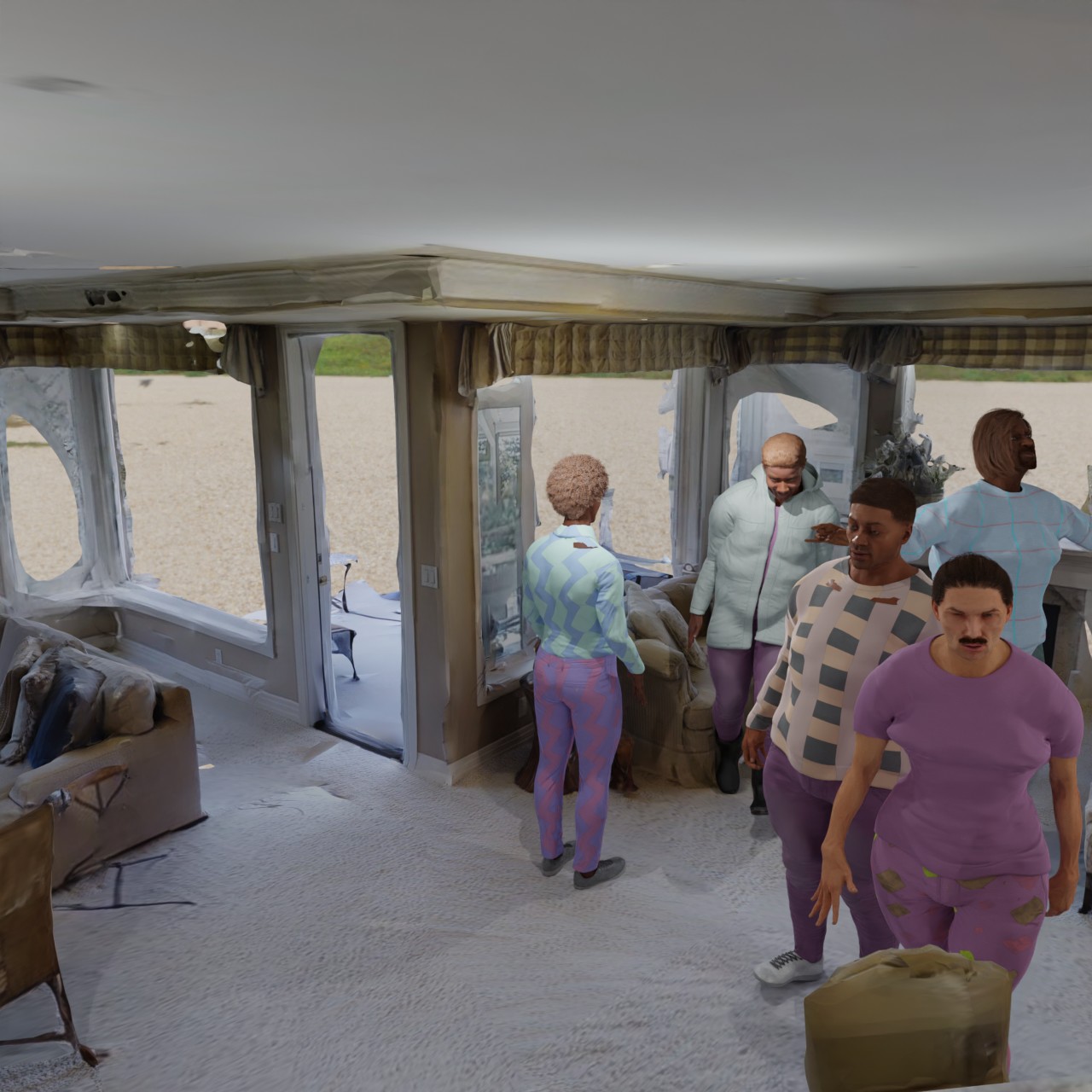}
    \includegraphics[width=0.4\linewidth]{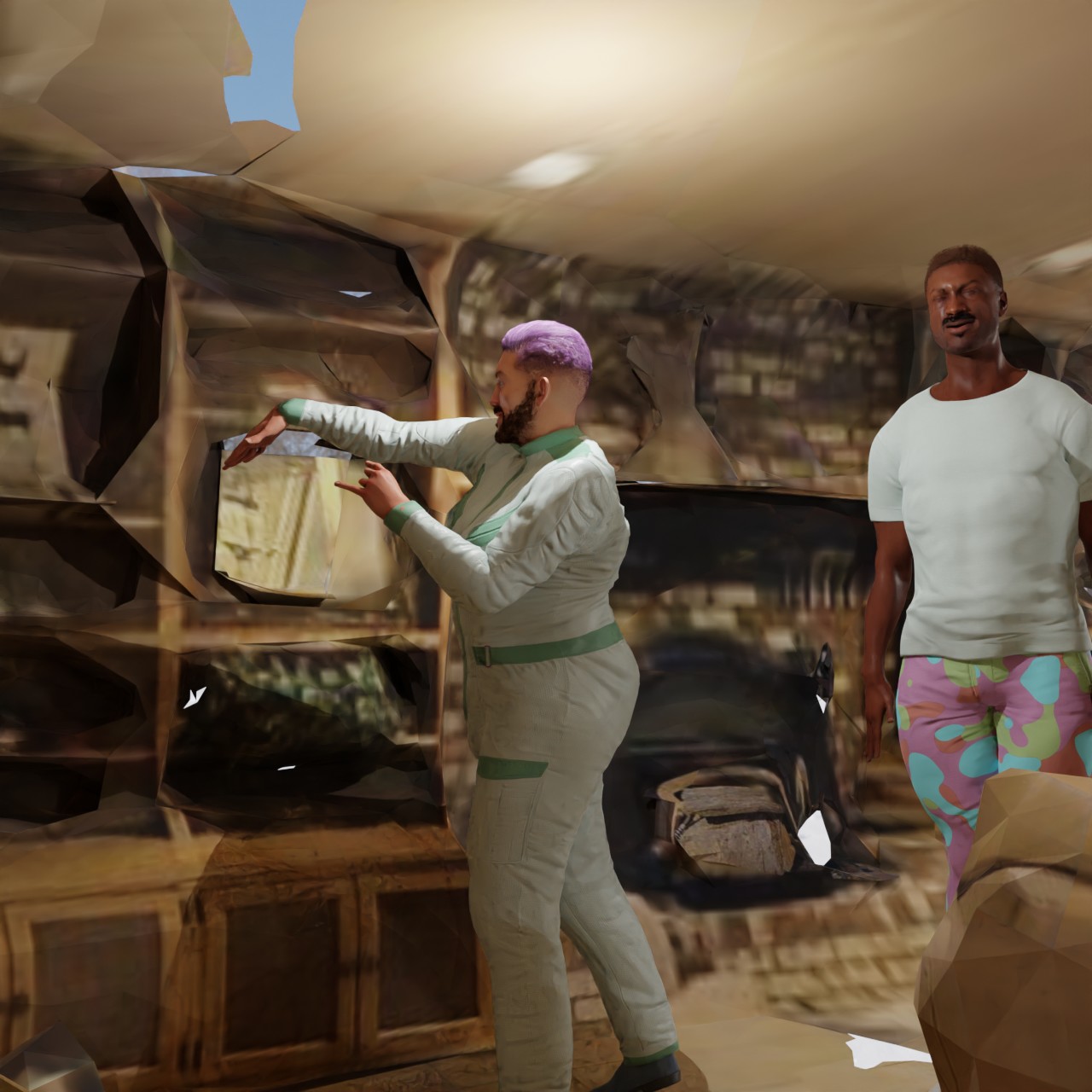}
    \vspace{-0.2cm}
    \caption{\textbf{Examples of synthetic renderings used in our training.}}
    \label{fig:our_synthetic_data}
\end{figure}

\begin{figure}
    \centering
    \includegraphics[width=\linewidth]{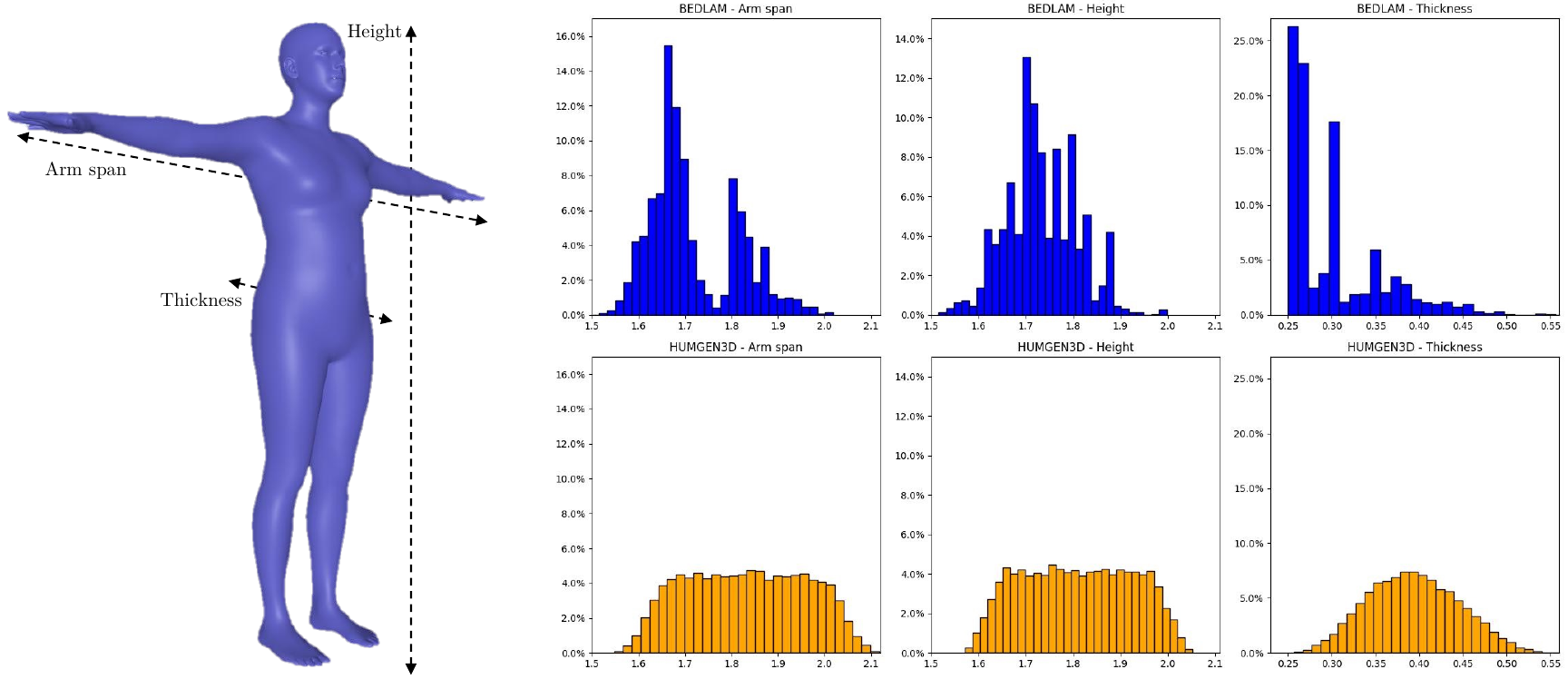}
    \vspace{-0.6cm}
    \caption{
    \textbf{Body shape statistics} on BEDLAM~\cite{black2023bedlam} and our synthetic data.
    } 
    \label{fig:data_variability}
\end{figure}

Fig.~\ref{fig:our_synthetic_data} illustrates the additional synthetic data generated to train our method. The images were rendered using Blender~\cite{blender}.
We created a collection of 3D scenes, each comprising a reconstructed indoor environment, an environment map for background and outdoor lighting, human characters, additional indoor light sources and cameras for rendering.
These components were procedurally selected and combined to enhance the realism of the scenes.
Specifically, we used scene meshes from Matterport3D~\cite{matterport3d}, Gibson~\cite{gibsonenv} and Habitat~\cite{habitat}, along with environment maps from PolyHaven~\cite{polyhaven}.
The characters were generated using HumGen3D~\cite{humgen3d}, a human generator plug-in for Blender.
Our synthetic data features a body shape distribution with a thicker tail than BEDLAM for increased diversity, as illustrated in Fig.~\ref{fig:data_variability}.

\subsection{Matching}
\paragraph{Matching additional inputs}
To perform experiments with external body shape (resp.\ distance) inputs, we associate to each prediction the shape (resp.\ distance) of the closest ground truth annotation, according to the 2D distance between their reference keypoints.

\paragraph{Evaluation}
For evaluation, we associate each ground truth mesh with the closest prediction, according to their PA-PJE distance.

\end{document}